\documentclass[lettersize,journal]{IEEEtran}
\usepackage{amsmath,amsfonts,amssymb}
\usepackage{algorithmic}
\usepackage{algorithm}
\usepackage{booktabs}
\usepackage{multicol}
\usepackage{multirow}
\usepackage{booktabs}
\usepackage{enumerate}
\usepackage{xcolor}
\usepackage{colortbl}
\usepackage{tabularray}
\usepackage{microtype}
\usepackage[utf8]{inputenc}
\usepackage[T1]{fontenc}
\UseTblrLibrary{booktabs}
\usepackage{tabularx}
\usepackage{hhline}
\usepackage{paralist}
\usepackage{marvosym}
\usepackage{array}
\usepackage[caption=false,font=scriptsize,labelfont=sf,textfont=sf]{subfig}
\usepackage{textcomp}
\usepackage{stfloats}
\usepackage{soul}
\usepackage{url}
\usepackage{verbatim}
\usepackage{graphicx}
\usepackage{hhline}
\usepackage{cite}
\usepackage[colorlinks,
linkcolor=black,
anchorcolor=black,
citecolor=black]{hyperref}
\newcommand{\tabincell}[2]{\begin{tabular}{@{}#1@{}}#2\end{tabular}}

\def\eg{\emph{e.g.}}

\def\etal{{\em et al.~}}

\begin{document}

\title{Learning Dynamic Local Context Representations for Infrared Small Target Detection}

\author{Guoyi~Zhang,~Guangsheng~Xu,~Han~Wang,~Siyang~Chen,~Yunxiao~Shan,~and~Xiaohu~Zhang
	\thanks{Manuscript received xxx, xxx; revised xxx, xxx. \emph{(corresponding authors: Yunxiao Shan and Xiaohu Zhang.)}}
	\thanks{Guoyi~Zhang, Guangsheng~Xu, Han~Wang, Siyang~Chen, and Xiaohu~Zhang are with the School of Aeronautics and Astronautics, Sun Yat-sen University, Shenzhen 518107, Guangdong, China.(email: zhanggy57@mail2.sysu.edu.cn;xugsh6@mail2.sysu.edu.cn;wangh737@mail2\\.sysu.edu.cn;chensy253@mail2.sysu.edu.cn;zhangxiaohu@mail.sysu.edu.cn)}
	\thanks{Yunxiao Shan is with the School of Artificial Intelligence, Sun Yat-sen University, Zhuhai, Guangdong 519082, China, also with the Southern Marine Science and Engineering Guangdong Laboratory, Sun Yat-sen University,
	Zhuhai, Guangdong 519082, China, also with Shenzhen Institute, Sun Yat-sen University, Shenzhen 510275, China, and also with the Guangdong Key Laboratory of Big Data Analysis and Processing, SunYat-sen University, Guangzhou 510006, China (e-mail: shanyx@mail.sysu.edu.cn).}
}

\markboth{Journal of \LaTeX\ Class Files,~Vol.~14, No.~8, August~2021}%
{Shell \MakeLowercase{\textit{et al.}}: A Sample Article Using IEEEtran.cls for IEEE Journals}


\maketitle

\begin{abstract}
Infrared small target detection (ISTD) is challenging due to complex backgrounds, low signal-to-clutter ratios, and varying target sizes and shapes. Effective detection relies on capturing local contextual information at the appropriate scale. However, small-kernel CNNs have limited receptive fields, leading to false alarms, while transformer models, with global receptive fields, often treat small targets as noise, resulting in miss-detections. Hybrid models struggle to bridge the semantic gap between CNNs and transformers, causing high complexity.
To address these challenges, we propose LCRNet, a novel method that learns dynamic local context representations for ISTD. The model consists of three components: (1) C2FBlock, inspired by PDE solvers, for efficient small target information capture; (2) DLC-Attention, a large-kernel attention mechanism that dynamically builds context and reduces feature redundancy; and (3) HLKConv, a hierarchical convolution operator based on large-kernel decomposition that preserves sparsity and mitigates the drawbacks of dilated convolutions. Despite its simplicity, with only 1.65M parameters, LCRNet achieves state-of-the-art (SOTA) performance.
Experiments on multiple datasets, comparing LCRNet with 33 SOTA methods, demonstrate its superior performance and efficiency.
\end{abstract}


\section{Introduction}
\IEEEPARstart{T}{he} infrared small target detection (ISTD) is vital for infrared search and tracking systems \cite{FEAIRSTD}, enabling long-range target capture and strong anti-interference capabilities in applications like reconnaissance, maritime surveillance, and precision guidance \cite{zhao2022single,strickland2023infrared,yi2023spatial}. However, ISTD in cluttered backgrounds remains a long-standing challenge due to the lack of distinctive features \cite{SeRankDet}, low signal-to-clutter ratio \cite{liu2023infrared}, and varying target shapes \cite{ISNet, SRNet, CSRNet}. To address these challenges, model-driven approaches were initially proposed. However, these methods rely on specific priors for the foreground, background, or both \cite{chen2022local}, and lack the ability to capture semantic information, leading to poor performance and limited generalization \cite{liu2023infrared}.

\begin{figure}[!t]
	\centering
	\includegraphics[width=\linewidth]{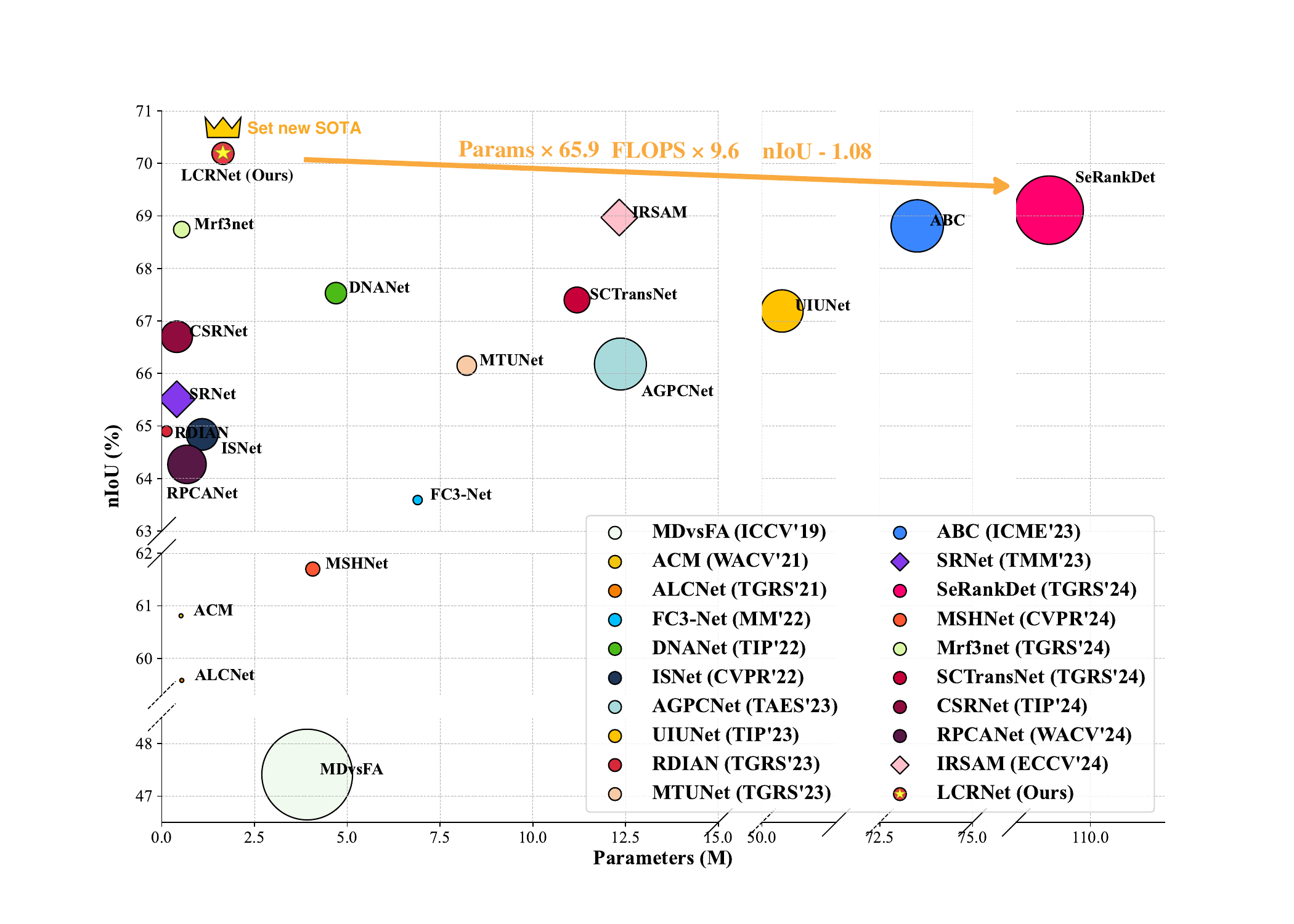}
	\caption{Comparison of the proposed LCRNet with other  data-driven methods on the IRSTD-1k dataset \cite{ISNet}. The area of the colorful circles represents the number of FLOPs, while the diamonds indicate cases where FLOPs are unknown. Our LCRNet achieves a remarkable balance between computational efficiency and detection performance, setting a new SOTA.}
	\label{fig:Compare}   
\end{figure}

In recent years, data-driven approaches have achieved remarkable performance and generalization, emerging as the dominant paradigm in the field, in contrast to model-driven approaches \cite{MSHNet}. Early methods \cite{DNANet, UIUNet} based on small-kernel CNNs have been proposed, incorporating priors such as contrast \cite{ALCNet, Dai_2021_WACV, RDIAN} and shape-bias \cite{ISNet,SRNet,CSRNet}  to enhance ISTD. The introduction of Transformer architectures \cite{liu2023infrared} has revitalized data-driven ISTD \cite{MTUNet, ABC, SeRankDet, SCTransNet}, leveraging self-attention mechanisms to capture long-range dependencies between features, thereby improving target-background discrimination and robustness in complex scenes. Recently, some work \cite{IRSAM} has focused on fine-tuning vision foundation models \cite{SAM}, which has improved segmentation performance for larger targets in the ISTD task. However, existing methods have not fully exploited the potential of data-driven approaches. As illustrated in Fig. \ref{fig:Compare}, these methods either underperform or require significantly higher computational resources—sometimes by orders of magnitude—just to achieve marginal improvements in detection accuracy. These limitations severely hinder the practical applicability of data-driven methods in ISTD.

\begin{figure}[!t]
	\centering
	\subfloat[]{
		\includegraphics[width=\linewidth]{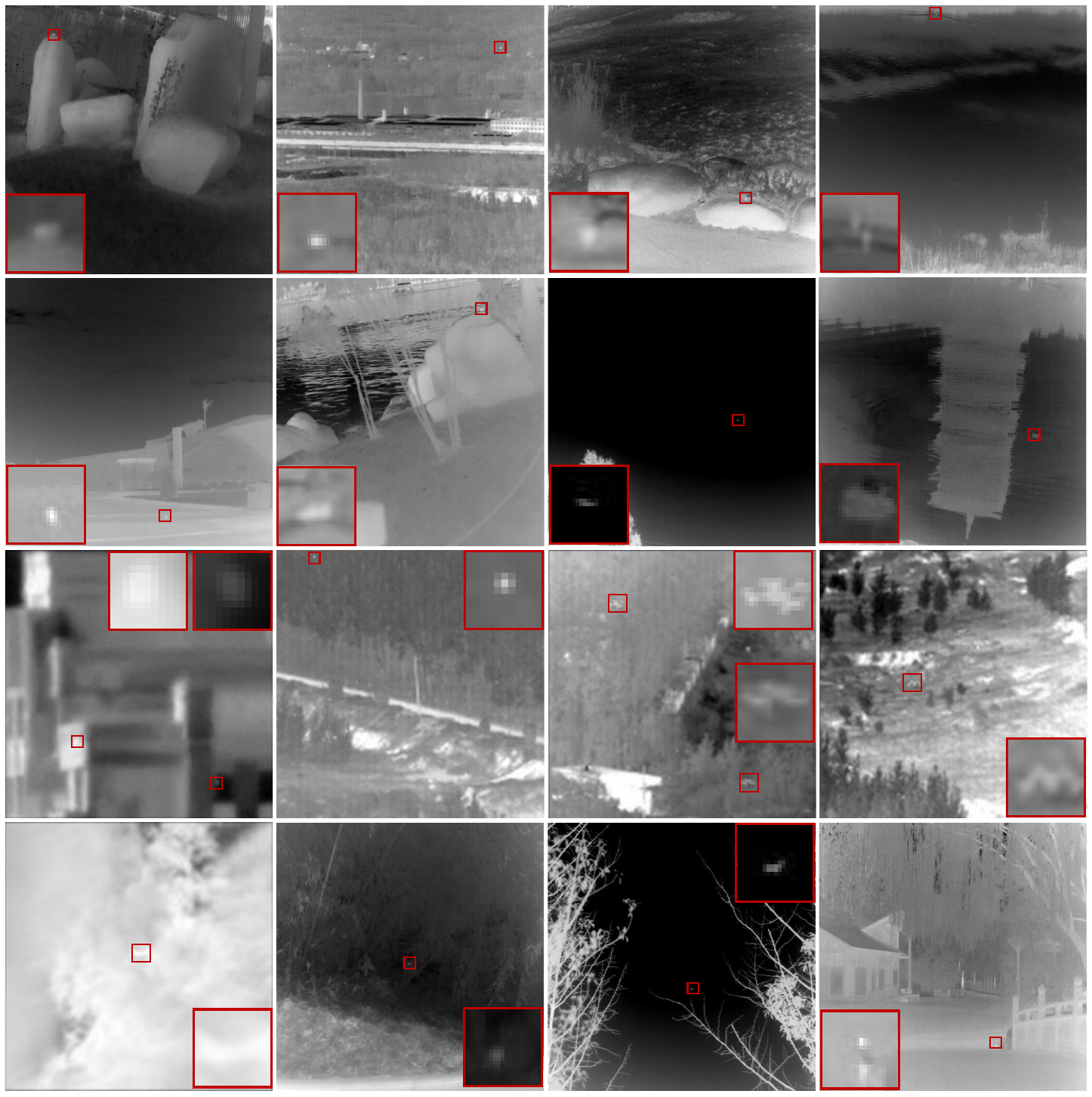}
		\label{fig:Local}
	}
	\hfill
	\subfloat[]{
		\includegraphics[width=\linewidth]{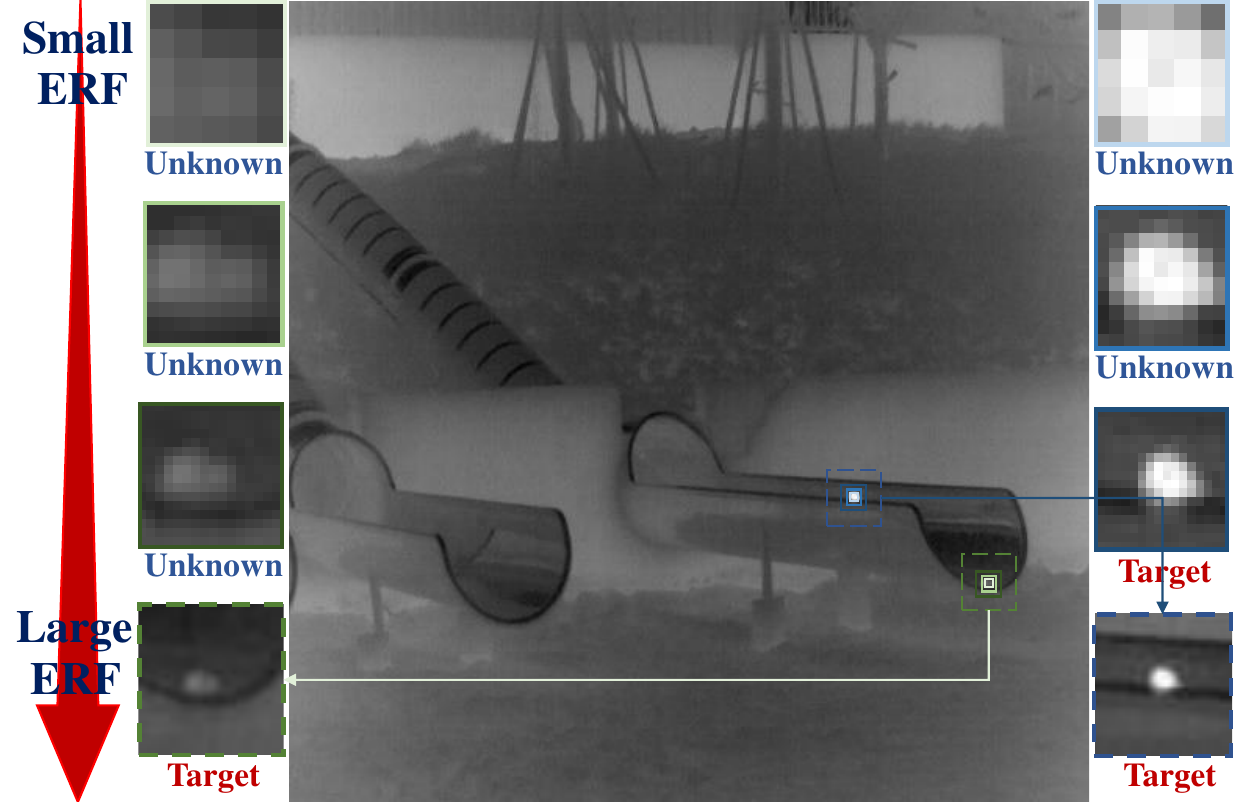}
		\label{fig:Priors}
	}
	\caption{Two key priors for infrared small target detection. (a) Small target information is confined to local regions of the image, and global observation often leads to missed detection of small targets. (b) Small target detection requires contextual information, where targets with varying shapes, sizes, and SCRs need different local contextual scales for accurate detection. As the context size increases, pixel-level details of the target are lost. Here, ERF denotes the size of the effective receptive field \cite{luo2016understanding}.}
	\label{fig:main}
\end{figure}

The issues with existing methods arise from their neglect of the unique characteristics inherent in the ISTD task. As shown in Fig. \ref{fig:Local}, ISTD involves complex and variable backgrounds (\eg, urban environments, sky, ocean), with foreground targets that are not specific to any particular category (\eg, person, ship, drone) \cite{ISNet}. Additionally, targets exhibit different properties at varying scales (\eg, small targets often resemble Gaussian distributions \cite{GRSL2014ILCM}, while larger targets exhibit category-specific shapes). Infrared images typically lack color and texture information, and are subject to strong noise and background clutter, resulting in low contrast \cite{gao2013infrared}. These challenges limit the applicability of simple contrast priors, with shape bias being effective only for larger targets. Moreover, the global receptive field of transformer models often leads to small targets being misinterpreted as random noise, causing miss-detections \cite{ABC}.

In fact, the ISTD task is characterized by two key priors: (1) \textbf{The locality of target information}. The number of target pixels is much smaller than that of background pixels, resulting in an extreme imbalance between target and background in the image. As shown in Fig. \ref{fig:Local}, small target information is confined to localized regions of the image, with little correlation to distant background information. (2) \textbf{The need for dynamic contextual information}. Small targets often lack distinguishable features, and their detection relies heavily on contextual information. As shown in Fig. \ref{fig:Priors}, small target detection requires varying contextual scales, where targets of different shapes, sizes, and SCRs necessitate different local contexts for accurate detection. However, as the context size increases, the pixel-level details of the target are lost.

To this end, we focus on the two key priors shown in Fig. \ref{fig:main}, and propose learning dynamic local context representations to facilitate ISTD. Specifically, the forward inference in the residual structure can be interpreted as solving differential equations \cite{he2019ode}, progressively refining the signal to accurately capture infrared small targets. Efficient PDE solvers, such as multigrid methods \cite{10061442}, iteratively reduce high- and low-frequency errors, yielding an accurate solution. Based on this analogy, we propose the coarse-to-fine convolution block (C2FBlock) to effectively capture small target information.
Next, we introduce the Dynamic Local Context Attention Mechanism (DLC-Attention), a large-kernel attention mechanism tailored for ISTD. This mechanism dynamically constructs contextual information at the appropriate scale for different inputs and reduces feature redundancy by grouping feature maps and aggregating information within each group.
Finally, we propose HLKConv, a hierarchical convolution operator based on large-kernel decomposition \cite{guo2023visual}. It preserves the large receptive field while maintaining sparse connectivity to optimize training. Moreover, HLKConv mitigates the negative effects of dilated convolutions on feature map holes with virtually no additional computational cost.

We summarize the main contributions of the paper as follows:
\begin{itemize}
	\item  We identify two key priors in the ISTD task and propose, for the first time, an end-to-end approach that enhances infrared small target detection by learning dynamic local context representations.
	\item Inspired by the multigrid method, we propose the C2FBlock to effectively capture small target information.
	\item We propose DLC-Attention, to the best of our knowledge, the first approach in this field to explore the dynamic construction of scale-aware and shape-sensitive local context representations, while reducing feature map channel redundancy via a large-kernel attention mechanism.
	\item  We propose HLKConv, a hierarchical convolution operator based on large-kernel decomposition \cite{guo2023visual}, which preserves sparsity and mitigates the drawbacks of dilated convolutions.
	\item  Extensive experiments on multiple datasets, comparing our approach with 33 state-of-the-art methods, show its superior performance and efficiency.
\end{itemize}

The remainder of this paper is organized as follows: In
Section \ref{Section:Related_Work}, related works are briefly reviewed. In Section \ref{Section:ProposedMethod},
we present the proposed method in detail. In Section \ref{Section:Experiment}, the
experimental results are given and discussed. Conclusions are
drawn in Section \ref{Section:Conclusion}.

\section{Related Work} \label{Section:Related_Work}
We briefly review related works from two aspects: infrared small target detection, and large-kernel networks.
\subsection{Infrared Small Target Detection}
Traditional infrared small target detection methods rely on specific priors for the foreground, background, or both. These can be broadly classified into Background Suppression Methods \cite{bai2010analysis,GRSL2014ILCM,PR16MPCM,deng2016small,GRSL2018RLCM,qin2019infrared,han2019local,qiu2022global} and Low-rank and Sparse Decomposition techniques \cite{gao2013infrared,dai2017non,dai2017reweighted,zhang2018infrared,zhang2019infrared1,zhang2019infrared}. However, their dependence on scene-specific priors limits generalization and makes them less effective in complex scenarios due to the lack of semantic understanding.

Data-driven approaches have become the mainstream, demonstrating superior performance \cite{MSHNet}. DNANet \cite{DNANet} integrates multi-scale information through dense connections, effectively distinguishing targets from the background. UIUNet \cite{UIUNet} employs nested U-shaped subnetworks for more refined target-background separation.
AGPCNet \cite{AGPCNet} utilizes a non-local attention mechanism to incorporate contextual information for target detection; however, large-scale context can often treat small targets as noise and discard them. ISNet \cite{ISNet}, SRNet \cite{SRNet}, and CSRNet \cite{CSRNet} improve detection performance by introducing shape-bias of the target, but these priors are only effective for large-scale targets, limiting the model's adaptability and requiring fine-grained annotations in the dataset. Gao \etal \cite{liu2023infrared} introduced Transformer architectures into infrared small target detection, bringing new vitality to the task. Transformer-based models demonstrate stronger robustness to noise interference \cite{SeRankDet, SCTransNet, ABC}. However, these methods are limited by the high computational cost of the self-attention mechanism, making them difficult to apply to high-resolution feature maps where small targets, represented by low-level semantic information, predominantly reside \cite{li2023ilnet}. Moreover, the semantic gap \cite{huang2023large} between Transformer and CNN in hybrid architectures necessitates additional, carefully designed feature fusion models \cite{MTUNet}, thereby increasing overall complexity. Additionally, the low inductive bias of Transformers limits their applicability in infrared small target detection tasks, especially since these datasets often contain very few samples and require complex data augmentation strategies. Recent works \cite{RPCANet} have attempted to introduce deep expansion approaches for ISTD tasks, but lack suitable model designs, and although they offer some interpretability, their performance remains suboptimal. IRSAM \cite{IRSAM} attempts to explore SAM \cite{SAM} fine-tuning transfer methods tailored for ISTD tasks, which perform well for large targets but underperform for small targets.

Unlike previous works, we focus on two key priors of the ISTD task and explore the use of a large-kernel attention mechanism specifically suited for ISTD, as well as the introduction of C2FBlock, inspired by PDE solvers. Notably, our goal is not to achieve state-of-the-art performance, but rather to propose a simple yet effective model that validates the efficacy of incorporating these key priors. We deliberately forgo complex data augmentation strategies and tricks like deep supervision, which have been widely proven effective in ISTD tasks, to focus on the core aspects of the model's design.
\subsection{Large Kernel Networks}
Recent studies \cite{guo2023visual,ding2022scaling,ding2024unireplknet,hou2024conv2former,li2023moganet} have demonstrated that convolutional networks with large receptive fields can rival transformer-based models in performance. To address the computational complexity of large-kernel convolutions, various works have explored structural reparameterization techniques, reducing their complexity from 
$\mathcal{O}(N^2)$ \cite{guo2023visual} to $\mathcal{O}(NlogN)$ \cite{finder2025wavelet}, 
$\mathcal{O}(N)$ \cite{lau2024large}, and even $\mathcal{O}(log N)$ \cite{chen2024pelk}, ensuring sparse operators for smoother training optimization \cite{ding2022scaling}. Large-kernel convolutions have been applied across diverse tasks, including general object detection in SegNeXt \cite{guo2022segnext}, remote sensing target detection in LSKNet \cite{li2024lsknet,li2023large}, and medical image processing \cite{azad2024beyond}. In the realm of visual recognition, Conv2Former \cite{hou2024conv2former} demonstrated that large-kernel attention mechanisms outperform non-attention-based large convolutions, while UniRepLKNet \cite{ding2024unireplknet} extended large-kernel convolutions to multimodal tasks, achieving state-of-the-art performance and showing no inherent superiority between CNN and Transformer architectures. More recently, Lin et al. \cite{CSRNet,SRNet} explored the influence of shape bias introduced by large-kernel convolutions in ISTD. However, the exploration of large-kernel convolutions in ISTD remains limited. Specifically, shape bias priors \cite{CSRNet,SRNet,ISNet} tend to perform well only for larger targets, and the introduction of inappropriate priors restricts the model's generalization and adaptability.

Unlike the aforementioned works, we focus on designing an appropriate large-kernel attention mechanism to capture two key priors in the ISTD task, as shown in Fig. \ref{fig:main}, thereby enhancing the discriminability between infrared small targets and the background. Specifically, we propose DLC-Attention, which dynamically constructs suitable local context for different inputs and reduces redundancy in the feature map's channel dimensions. In contrast to the most similar work, LSKNet \cite{li2024lsknet,li2023large}, DLC-Attention does not adopt a hierarchical structure at a macro level, allowing for better adaptation to modern GPUs. Moreover, by employing aggregation and sampling strategies, we achieve broader contextual allocation with reduced computational cost. We also address redundancy in the feature map's channel dimensions by implementing group-based suppression. Finally, we introduce a linear layer and nonlinear activation function within the value matrix to enable higher-order interactions \cite{rao2022hornet}, thereby enhancing the expressive power of the model. Additionally, we design the HLKConv operator to ensure a large receptive field while maintaining sparse connectivity, mitigating the negative effects of dilated convolutions during the structural reparameterization process \cite{guo2023visual, lau2024large, li2023moganet}, thereby better adapting to the fine-grained nature of small target detection.

\section{Proposed Method} \label{Section:ProposedMethod}
\subsection{Overview}
\begin{figure*}[!t]
	\centering
	\includegraphics[width=1\linewidth]{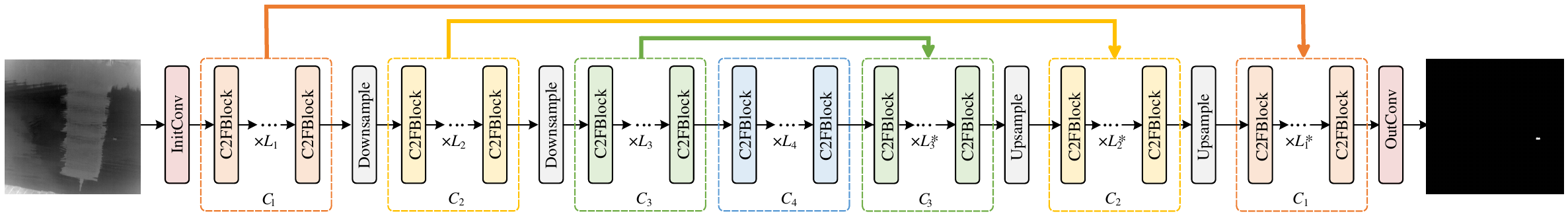}
	\caption{The overall architecture of LCRNet follows the typical U-Net structure \cite{ho2020denoising, nichol2021improved, lugmayr2022repaint}, with stacked C2FBlocks. The InitConv and OutConv are two 3×3 convolutions: InitConv increases the number of channels from 1 to $C_1$, while OutConv reduces the channels from $C_1$ back to 1. Specifically, $L^{*}_i$ = $L_i + 1$, with $L_1 = L_2 = L_3 = L_4 = 3$ and channel sizes $C_1 = 16$, $C_2 = 32$, $C_3 = 64$, and $C_4 = 64$, The total number of parameters is 1.65M, and the FLOPs is 59.3G.}
	\label{fig:OverallofAMRI-Net}
\end{figure*}

As shown in Fig. \ref{fig:OverallofAMRI-Net}, the overall architecture of the proposed method follows the widely adopted U-Net structure \cite{ho2020denoising, nichol2021improved, lugmayr2022repaint} used in recent years. LCRNet is composed of stacked proposed C2FBlocks, with the core components of C2FBlock being the proposed DLC-Attention and HLKConv.

The model takes an image with a size of \(1 \times H \times W\) as input. It is first processed by a 3\( \times \)3 convolution (InitConv) to increase the feature map’s dimensionality to \(C_1\), and then undergoes a series of C2FBlocks for processing at different resolutions. Finally, a 3\( \times \)3 convolution (OutConv) reduces the feature map’s dimensionality from \(C_1\) to 1, producing a confidence map. Consistent with previous works \cite{UIUNet,ABC,SeRankDet}, we apply a fixed threshold of 0.5 to segment the confidence map into the final output.

\subsection{Coarse-to-fine convolution Block}
\begin{figure}[!t]
	\centering
	\includegraphics[width=1\linewidth]{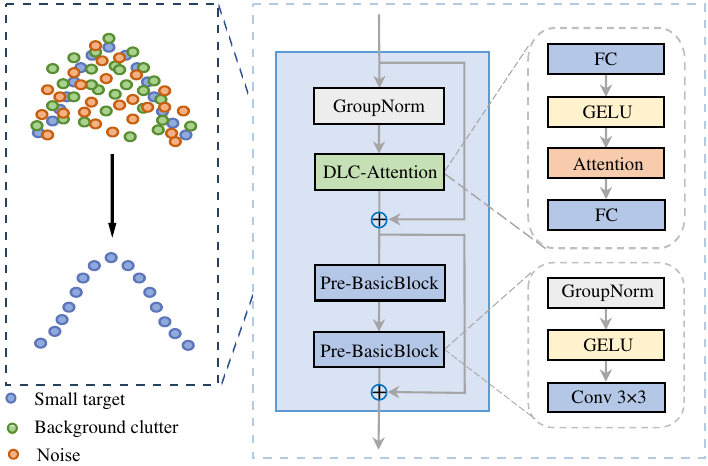}
	\caption{The overall architecture of the proposed Coarse-to-fine Convolution Block (C2FBlock) simulates the iterative process of a multigrid method \cite{10061442}. C2FBlock effectively distinguishes between spatial and frequency-domain distributions of similar targets, background clutter, and noise.}
	\label{fig:OverallofConvBlock}
\end{figure}
As shown in Fig. \ref{fig:OverallofConvBlock}, the proposed Coarse-to-fine Convolution Block (C2FBlock) is a fundamental component of LCRNet. 

\noindent\textbf{Motivation}. First, MetaFormer \cite{yu2022metaformer,yu2023metaformer} demonstrates that modern module designs can guarantee the lower bound of model performance. However, the original MetaFormer-style modules are not well-suited for fine-grained tasks \cite{shi2024transnext}. Second, the forward inference in the residual structure can be interpreted as solving differential equations \cite{he2019ode}, progressively refining the signal to accurately capture infrared small targets. This suggests that we can adapt MetaFormer \cite{yu2022metaformer,yu2023metaformer} by considering the design of modules from the perspective of efficient and precise differential equation solvers for ISTD. Specifically, efficient PDE solvers, such as multigrid methods \cite{10061442}, iteratively reduce both high- and low-frequency errors, yielding an accurate solution. This insight inspires the design of modules based on the iterative process of multigrid methods. One potential approach is to combine different effective receptive fields, as small receptive fields function like high-pass filters, eliminating low-frequency errors \cite{ABC}, while large receptive fields act as low-pass filters, removing high-frequency errors \cite{10595430}. Furthermore, the pre-act residual block \cite{huang2023revisiting} can be viewed as a special case of a multigrid iteration process \cite{he2019mgnet}. Therefore, we combine the proposed DLC-Attention with the pre-act residual block to form a coarse-to-fine convolution module (C2FBlock), better capturing the small target information.
For an input feature map $\mathbf{X} \in\mathbb{R}^{W\times H \times C}$,  the processing procedure can be expressed as:
\begin{equation}
	\mathbf{X}^{\prime}=\mathbf{X}+\lambda_1\text{DLC-Attention}\left(\mathrm{Norm}_1(\mathbf{X})\right),
\end{equation}
\begin{equation}
	\mathbf{X}^{\prime\prime}=\mathbf{X}^{\prime}+\lambda_2\text{pre-BasicBlock}\left(\text{pre-BasicBlock}\left(\mathbf{X}^{\prime}\right)\right).
\end{equation}
where $\mathrm{Norm}_1(\cdot)$ denotes Group Normalization \cite{wu2018group}, and pre-BasicBlock \cite{huang2023revisiting} is a component of pre-act ResBlock \cite{huang2023revisiting}. $\lambda_1$ and $\lambda_2$ are learnable parameters that define the residual weighting in the LayerScale \cite{touvron2021going} mechanism.

\noindent\textbf{Advantages.} Compared to modern designs like MetaFormer \cite{yu2022metaformer, yu2023metaformer}, the proposed approach offers two key advantages: (1) The coarse-to-fine iterative process, inspired by the multigrid method \cite{10061442}, enables more precise capture of small target information; (2) Unlike FFNs \cite{xie2021segformer, wang2021pyramid} in modern architectures, which typically expand the channel dimension by a factor of four, C2FBlock maintains consistent input and output dimensions, reducing memory consumption and optimizing compatibility with modern GPUs. In contrast to replacing FFNs with conventional ResBlocks \cite{he2016deep}, the proposed C2FBlock simulates the characteristics of the multigrid method \cite{10061442} to capture small target information, leading to significant performance improvements without incurring any additional computational cost.

\subsection{Dynamic Local Context Attention Mechanism}
\begin{figure*}[!t]
	\centering
	\includegraphics[width=1\linewidth]{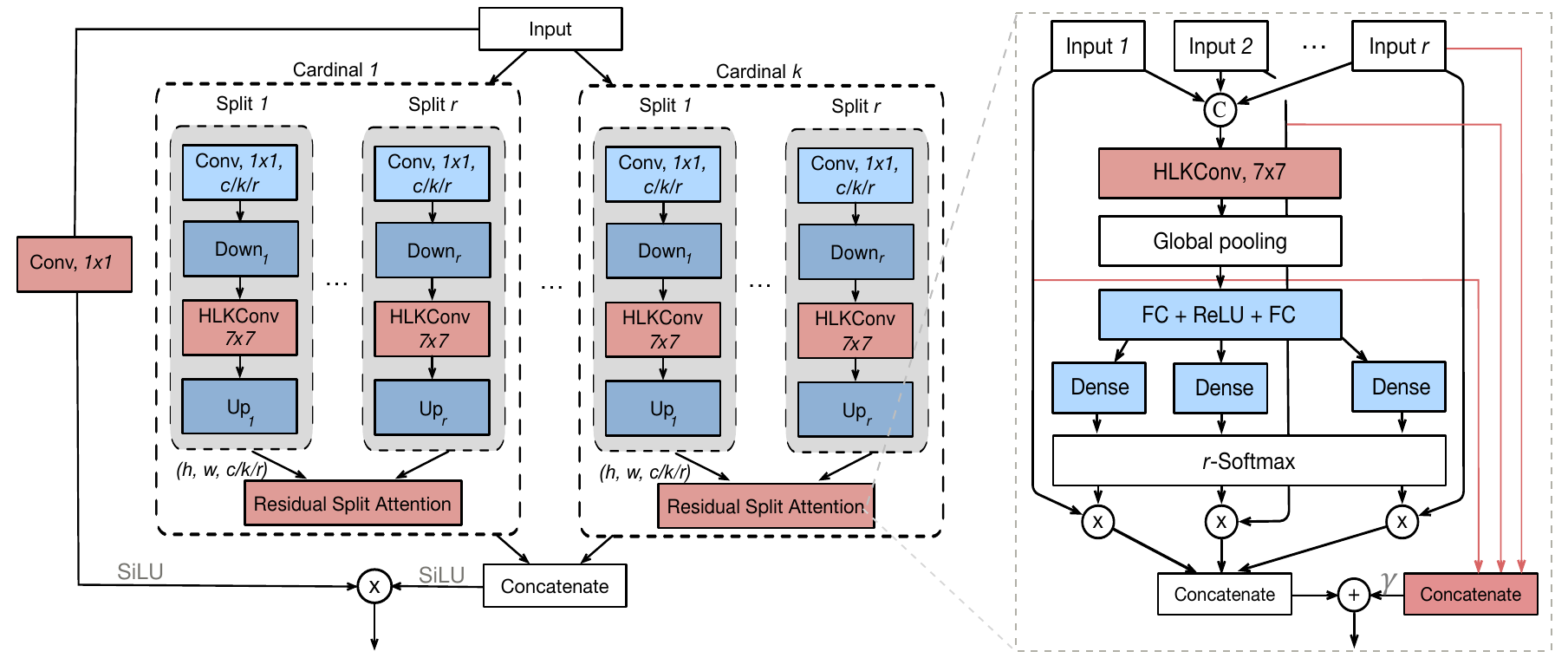}
	\caption{Overall architecture of the proposed attention. For simplicity, we show proposed attention in cardinality-major view (the featuremap groups with same
	cardinal group index reside next to each other). We use radix-major \cite{zhang2022resnest} in the real implementation, which can be modularized and accelerated by group convolution and standard CNN layers.}
	\label{fig:OverallofAttention}
\end{figure*}
Our DLC-Attention shares a similar overall architecture with recent large-kernel attention mechanisms \cite{guo2023visual, li2024lsknet, hou2024conv2former, li2023moganet}, as shown in Fig. \ref{fig:OverallofConvBlock}. The key difference lies in the composition of the attention mechanism itself, which is detailed in Fig. \ref{fig:OverallofAttention}. It consists of feature map groups, where each group captures multi-scale, long-range features. Within each group, the residual split-attention operation dynamically allocates intra-group weights to achieve dynamic receptive field allocation. Additionally, the feature map splitting effectively suppresses redundancy in the channel dimensions of the feature maps.

\noindent\textbf{Overview of the proposed attention.} For an input feature map $\mathbf{X} \in\mathbb{R}^{W\times H \times C}$, the attention mechanism first computes the similarity score matrix $\mathbf{A} \in\mathbb{R}^{W\times H \times C}$ and the value matrix $\mathbf{V} \in\mathbb{R}^{W\times H \times C}$. Then, it uses the Hadamard product to calculate the output $\mathbf{Z} \in\mathbb{R}^{W\times H \times C}$. The entire process is as follows:
\begin{equation}
	\mathbf{Z} = \text{SiLU}(\mathbf{A})\odot\text{SiLU}(\mathbf{V}),
\end{equation}
where $\text{SiLU}(\cdot)$ is  the sigmoid linear unit function \cite{hendrycks2016gaussian}. The computation of the value matrix $\mathbf{V}$ is relatively straightforward, as follows:
\begin{equation}
	\mathbf{V} = \mathbf{W}_1\mathbf{X}.
\end{equation}
where $\mathbf{W}_1$ is weight
matrix of linear layer. The similarity score matrix $\mathbf{A}$ is the core of the proposed attention mechanism, but its computation is more complex. A detailed discussion of this process follows.

\noindent\textbf{Featuremap group.} The successful experiences of architectures \cite{zhang2022resnest} demonstrate that grouping feature maps through convolution can reduce redundancy in the channel dimension, which is crucial for enhancing the expressive power of narrower networks. Therefore, in the computation of the similarity score matrix $\mathbf{A}$, the first step is to perform grouping, followed by applying a series of transformations \( \{\mathcal{T}_1, \mathcal{T}_2, ..., \mathcal{T}_K \} \) to each individual group. The intermediate representation of each group is then obtained as: 
\begin{equation}
	\mathbf{X}_i = \mathcal{T}_i(\mathbf{X}).
\end{equation}

\begin{figure}
	\centering
	\includegraphics[width=1\linewidth]{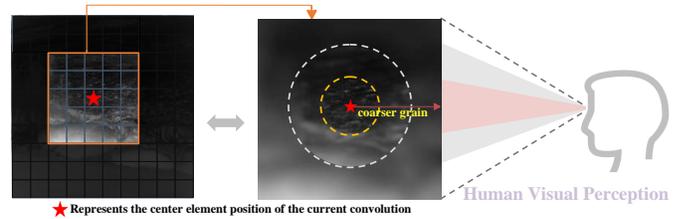}
	\caption{The multi-scale aggregation in each group (for \(r=3\)) facilitates target-level discrimination by leveraging large context information. Upsampling and downsampling help retain contextual attention while reducing the model's computational load, allowing for focused refinement of small regions.}
	\label{fig:biological_vision3}
\end{figure}

\noindent\textbf{Multi-scale aggregation in each group.} 
To introduce the ability for long-range multi-scale interaction and dynamic modeling into convolution, we adopt an aggregated convolution approach. In this method, the feature map is split along the channel dimension and computed in parallel across multiple branches. Each branch applies different degrees of downsampling to capture information at varying scales while reducing computational cost. Furthermore, each branch employs HLKConv to model long-range feature dependencies, promoting the separation of the target from background noise sources. 
The role of multi-scale aggregation is illustrated in Fig. \ref{fig:biological_vision3}. 
This aggregation method effectively simulates the human visual system's ability to perceive context while maintaining pixel-level focus.

\begin{equation}
	\left\{\mathbf{U}_1, ..., \mathbf{U}_r\right\} = \text{Split}(\mathbf{X}_i),1\leq i\leq K,
\end{equation}
\begin{equation}
	\hat{\mathbf{U}}_j=\uparrow_p\text{ (HLKConv}_{7\times7}(\downarrow_{\frac p{2^j}}(\mathbf{U}_j))),1\leq j\leq r.\label{Eq:DownandUp}
\end{equation} 
where $\uparrow_{p} (\cdot)$ represents upsampling features at a specific level to the original resolution
$p$  via nearest interpolation for fast implementation and $\downarrow\frac p{2^j} (\cdot)$ denotes maxpooling the input features to the size of $\frac p{2^j}$. 

\noindent It is worth noting that in this paper, we set \( r = 4 \). On one hand, some empirical evidence suggests that dividing the aggregated convolution into 4 parts yields optimal performance \cite{chen2023run}. On the other hand, a larger value of \( r \) would result in convolutional kernels that exceed the size of the feature map after multiple downsampling operations. In such cases, achieving optimal performance often requires complex data augmentation strategies and large-capacity datasets \cite{ding2022scaling,ding2024unireplknet}.

\noindent\textbf{Residual split attention.} To achieve dynamic allocation of the effective receptive field, we introduce an attention mechanism for the fusion of multi-scale information. Split attention \cite{li2019selective} is a suitable choice, but it differs from the original split attention in the following ways: (1) we first apply HLKConv in split attention to use large-scale convolutions to assess the importance of different branches in the spatial dimension; (2) we introduce a learnable temperature parameter \cite{hinton2015distilling} in r-softmax to adaptively modulate the sparsity of the output; (3) we incorporate residual connections to prevent the overconfidence of softmax \cite{xu2024eviprompt} and mitigate the competitive nature of softmax, which can suppress key features.

\noindent\textbf{Advantages.}
Compared to related works, the proposed attention mechanism offers several key advantages. Unlike Conv2Former \cite{hou2024conv2former} and VAN \cite{guo2023visual}, it dynamically allocates the effective receptive field to adapt to input variations while simultaneously reducing redundancy in the channel dimension. In contrast to ResNeSt \cite{zhang2022resnest}, the proposed mechanism not only adapts the receptive field dynamically but also enhances the model's expressive power through high-order interactions \cite{rao2022hornet} enabled by the Hadamard product \cite{ma2024rewrite}. While LSKNet \cite{li2023large,li2024lsknet} also allows for dynamic allocation of the receptive field, its serial operations reduce inference speed, and it lacks specialized optimizations for minimizing redundancy in the channel dimension. 

\subsection{HLKConv}
\begin{figure}[!t]
	\centering
	\includegraphics[width=\linewidth]{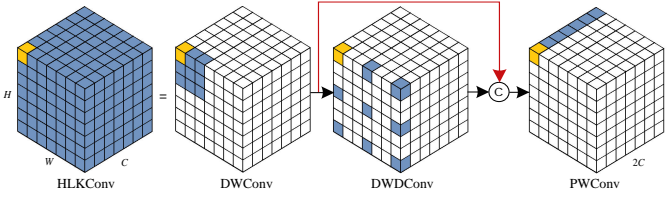}
	\caption{The overall architecture of the proposed HLKConv. It is based on the large-kernel decomposition principle \cite{guo2023visual} and incorporated into a hierarchical convolution approach to simultaneously capture both small and large effective receptive fields, while alleviating the feature map artifacts introduced by DWDConv convolutions.}
	\label{fig:OverallofLKConv}
\end{figure}
The proposed HLKConv, shown in Fig. \ref{fig:OverallofLKConv}, builds upon the large-kernel decomposition rule introduced by VAN \cite{guo2023visual}. The key distinction is that we incorporate a hierarchical convolution approach to mitigate the loss of feature map details caused by dilated convolutions. This is particularly crucial for small object detection, where preserving fine-grained details significantly enhances detection accuracy.

\noindent\textbf{Large-kernel decomposition rule.} A standard
convolution can be decomposed into three parts: a depth-wise convolution (DWConv), a depth-wise dilation convolution (DWDConv), and a pointwise convolution (1$\times$1 Conv) \cite{lau2024large}.
Specifically, a $K \times K$ convolution can be decomposed into 
a $\lceil \frac{K}{d} \rceil \times \lceil \frac{K}{d} \rceil$ depth-wise dilation convolution with dilation $d$, 
a $(2d-1) \times (2d-1)$ depth-wise convolution 
and a 1$\times$1 convolution \cite{guo2023visual}.

\begin{figure}
	\centering
	\includegraphics[width=1\linewidth]{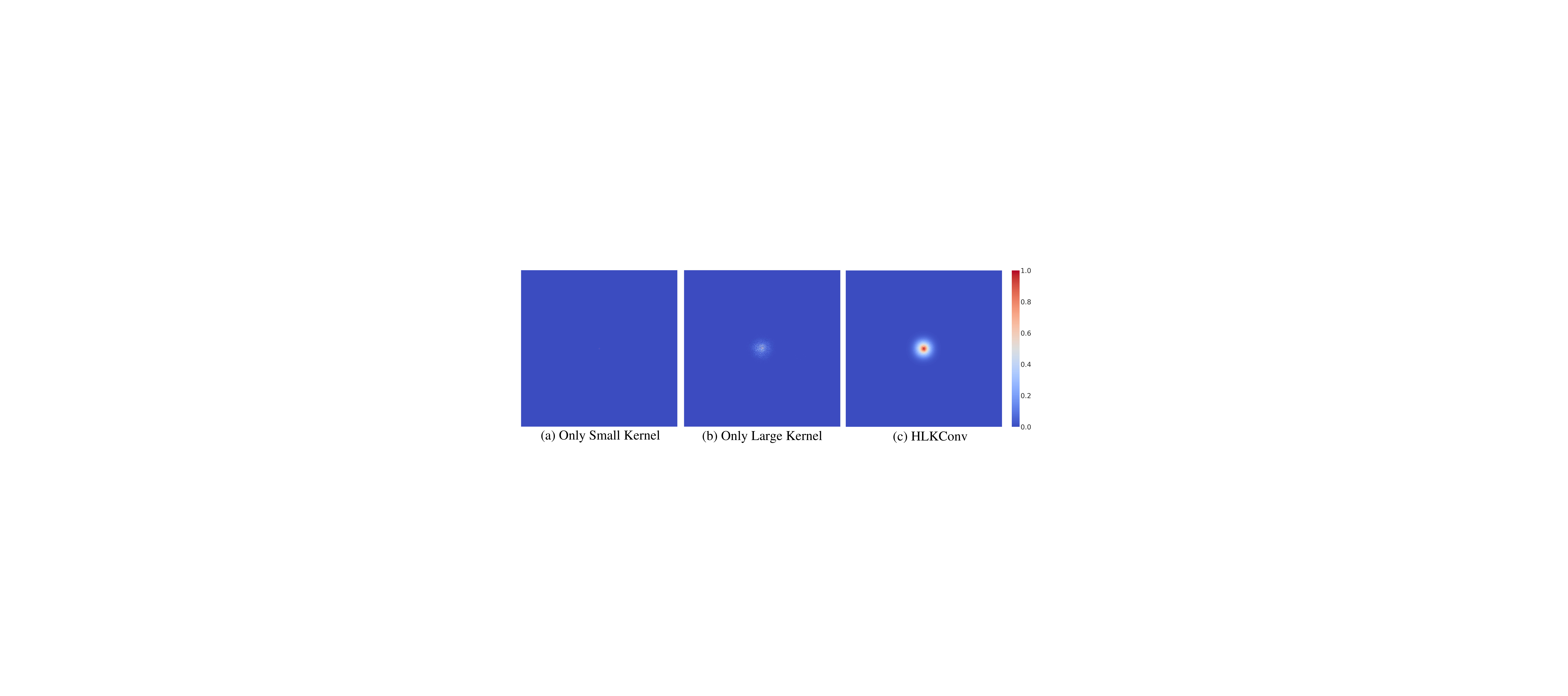}
	\caption{Compare the effective receptive fields after random initialization for LCRNet. It can be observed that, under the proposed HLKConv, the effective receptive field \cite{luo2016understanding} in the initial state exhibits stronger Foveal Visual characteristics \cite{shi2024transnext}, which is more conducive to fine-grained tasks.}
	\label{fig:ERFCompared}
\end{figure}

\noindent\textbf{The proposed HLKConv.} The large-kernel decomposition rule offers a low-cost reparameterization for large kernels but introduces several challenges. First, DWDConv generates block-like artifacts in the feature maps, harming fine-grained tasks \cite{wang2024multi}. Second, small objects within a large receptive field lack detailed information, impairing segmentation accuracy \cite{MRF3Net,SRNet,CSRNet}. Additionally, attention maps struggle to capture both large-area and small-region information simultaneously, though recent studies suggest that combining these two can significantly enhance performance and generalization \cite{xie2023revealing,shi2024transnext,li2023moganet}.

To address these issues, we introduce a hierarchical convolution approach that mitigates the loss of feature details and block-like artifacts from DWDConv, while capturing both local and long-range features. This method adapts the effective receptive field at different stages of the model, ensuring that it meets the varying feature capture requirements at each stage. For an input feature map $\mathbf{X} \in\mathbb{R}^{W\times H \times C}$, the computation process of HLKConv is as follows:
\begin{equation}
	\hat{\mathbf{X}} = \text{DWConv}_{\lceil \frac{K}{d} \rceil \times \lceil \frac{K}{d} \rceil}(\mathbf{X}),
\end{equation}
\begin{equation}
	\mathbf{X}^{\prime} = \text{Concat}(\hat{\mathbf{X}}, \text{DWDConv}_{(2d-1) \times (2d-1)}(\hat{\mathbf{X}})),
\end{equation}
\begin{equation}
	\mathbf{Y} = \text{Conv}_{1\times1}(\mathbf{X}^{\prime}).
\end{equation}
where $K$ is the size of the equivalent large kernel, $d$ is the dilation rate, and $\mathbf{Y} \in \mathbb{R}^{W \times H \times C}$ is the output after processing with HLKConv. Fig. \ref{fig:ERFCompared} shows the effective receptive fields \cite{luo2016understanding} of LCRNet after random initialization, comparing the influence of HLKConv, LKConv \cite{guo2023visual}, and small kernels. Notably, with HLKConv, LCRNet’s effective receptive field aligns more closely with Foveal Visual Perception \cite{shi2024transnext}, which is known to benefit fine-grained tasks.

\noindent\textbf{Complexity analysis.} We present the parameters and floating point operations (FLOPs) of our HLKConv. Bias is omitted in the computation process for simplifying format. We assume that the input and output features have same size $W \times H \times C$. The number of parameters $P(K, d)$ and FLOPs $F(K, d)$ can be denoted as:
\begin{equation}
	P(K, d) = C \times \left({\lceil \frac{K}{d} \rceil}^2 + (2d-1)^2\right) + 2C^2,
\end{equation}
\begin{equation}
	F(K, d) = P(K, d) \times W \times H.
\end{equation}
where $d$ denotes the dilation rate and $K$ is the kernel size. According to the formulas for FLOPs and parameters, the dilation saving ratio is consistent for both FLOPs and parameters.

\subsection{Loss Function}
To handle the class imbalance issue between targets and
backgrounds and focus more on small target regions \cite{FEAIRSTD}, the Soft Intersection of Union (Soft-IoU) loss \cite{SoftIoU} is adopted to calculate the distance between the confidence map and the ground truth, defined by:
\begin{equation}
	\mathcal{L} = 1 - \frac{\sum_{i,j}\mathbf{X}_{i,j}\cdot\mathbf{Y}_{i,j}}{\sum_{i,j}\mathbf{X}_{i,j}+\mathbf{Y}_{i,j}-\mathbf{X}_{i,j}\cdot\mathbf{Y}_{i,j}}
\end{equation}
where $\mathbf{X}$ is the confidence map, and the $\mathbf{Y}$  is the ground truth image.

\section{Experiment}\label{Section:Experiment}
\subsection{Experimental Setup}
\subsubsection{Datasets} We adopt the widely used IRSTD-1K \cite{ISNet}, SIRSTAUG \cite{AGPCNet}, and NUDT-SIRST \cite{DNANet} to evaluate the proposed method. These datasets, renowned for their diversity and complexity \cite{SeRankDet}, provide a rigorous testing ground for our proposed method, ensuring a comprehensive evaluation of its performance and generalization capabilities. 

\subsubsection{Evaluation Metrics} In this study, four widely used evaluation metrics including the probability of detection ($P_d$), false alarm rate ($F_a$), intersection over union (IoU) and normalized intersection over union (nIoU) \cite{Dai_2021_WACV}, are used for performance evaluation. Among them, the $P_d$ and $F_a$ are object-level metrics, which are defined as follows:
\begin{equation}
	P_d=\frac{\#\text{ number of true detections }}{\#\text{ number of real targets }},
\end{equation}
\begin{equation}
	F_a=\frac{\#\text{number of false detections }}{\#\text{ number of images }}.
\end{equation}
The IoU and nIoU \cite{Dai_2021_WACV} are pixel-level metrics, which are defined as follows:
\begin{equation}
	IoU=\frac{TP}{T+P-TP},
\end{equation}
\begin{equation}
	nIoU=\frac{1}{N}\sum_i^N\frac{TP(i)}{T(i)+P(i)-TP(i)}.
\end{equation}
where $N$ is the total number of samples, $T$, $P$ and $TP$ denote the number of ground truth, predicted positive pixels, and the number of true positive respectively. The IoU metric is heavily influenced by the segmentation accuracy of large objects \cite{Cheng_2021_CVPR}, with large-scale targets in infrared small target datasets contributing more to the score. In contrast, the nIoU metric \cite{Dai_2021_WACV}, designed for small target detection, balances scale impact and offers a fairer evaluation. Together, IoU and nIoU provide a comprehensive view of the algorithm's adaptability to targets of varying scales, with a greater emphasis on nIoU.

\subsubsection{Implementation Details} The proposed method is implemented in PyTorch 1.7.1 and accelerated with CUDA 11.2. The network is trained using the Adan optimizer \cite{ADAN} with an initial learning rate of 1e-3, betas (0.98, 0.92, 0.99), eps 1e-8, max grad norm 0, and weight decay 1e-4 on an NVIDIA A100 GPU. Learning rate decay follows the poly policy \cite{RPCANet}, with a batch size of 8 and a maximum of 400 epochs. To ensure reproducibility, no data augmentation is used, and the input is a single-channel image normalized by dividing grayscale values by 255.0.

\subsection{Comparison with State-of-the-Arts}
\begin{table*}
	\renewcommand\arraystretch{1.2}
	\footnotesize
	\centering
	\caption{Comparison with Other State-of-the-art methods on three datasets. \textbf{\textcolor{cyan}{Blue}}, \textbf{\textcolor{orange}{orange}}, and \textbf{\textcolor{green}{green}} represent the best, second-best, and third-best metrics in each column, respectively.}
	\label{tab:sota}
	\setlength{\tabcolsep}{3.5pt}
	\begin{tabular}{l|c|c|c|cccc|cccc|cccc}
		\noalign{\hrule height 1pt}
		\multirow{2}{*}{Method} & \multirow{2}{*}{Publish} & \multirow{2}{*}{Param} & \multirow{2}{*}{FLOPs} & \multicolumn{4}{c|}{IRSTD-1k}   & \multicolumn{4}{c|}{SIRSTAUG}   & \multicolumn{4}{c}{NUDT-SIRST} \\ \multicolumn{1}{l|}{} & \multicolumn{1}{l|}{}  &\multicolumn{1}{l|}{} & \multicolumn{1}{l|}{} 
		& IoU $\uparrow$ &  nIoU $\uparrow$ & $P_d$ $\uparrow$ & $F_a$ $\downarrow$ & IoU $\uparrow$ &  nIoU $\uparrow$ & $P_d$ $\uparrow$ & $F_a$ $\downarrow$ & IoU $\uparrow$ &  nIoU $\uparrow$ & $P_d$ $\uparrow$ & $F_a$ $\downarrow$ \\
		\noalign{\hrule height 1pt}
		\multicolumn{16}{l}{\textit{Background Suppression Methods}}  \\
		\hline
		NWMTH \cite{bai2010analysis}  & PR'10   & $\textemdash$ & $\textemdash$ & 18.94 & 16.91 & 49.26 & 21.72 & 18.71 & 16.74 & 62.71 & 32.46 & 11.72 & 10.26 & 52.71 & 46.81 \\
		ILCM \cite{GRSL2014ILCM}  & GRSL'14 & $\textemdash$ & $\textemdash$ & 13.05 & 11.46 & 43.55 & 47.32 & 5.25  & 6.44  & 26.27 & 88.43 & 12.84 & 13.69 & 27.62 & 43.21 \\
		MPCM \cite{PR16MPCM}  & RR'16   & $\textemdash$ & $\textemdash$ & 20.04 & 23.63 & 58.46 & 31.15 & 27.39 & 38.56 & 55.43 & 11.47 & 9.52  & 9.96  & 43.8  & 79.22 \\
		WLDM \cite{deng2016small}  & TGRS'23 & $\textemdash$ & $\textemdash$ & 8.35  & 10.34 & 54.14 & 14.31 & 8.63  & 11.07 & 16.58 & 64.32 & 11.04 & 21.03 & 12.69 & 78.26 \\
		RLCM \cite{GRSL2018RLCM}  & GRSL'18 & $\textemdash$ & $\textemdash$ & 16.92 & 21.03 & 70.36 & 67.6  & 8.4   & 11.9  & 70.53 & 31.44 & 18.37 & 18.11 & 79.51 & 67.13 \\
		FKRW \cite{qin2019infrared}  & TGRS'19 & $\textemdash$ & $\textemdash$ & 10.39 & 16.25 & 69.54 & 24.37 & 10.05 & 16.56 & 72.42 & 66.87 & 12.67 & 21.73 & 79.51 & 67.13 \\
		TLLCM \cite{han2019local} & GRSL'20 & $\textemdash$ & $\textemdash$ & 10.26 & 18.24 & 68.55 & 24.48 & 13.96 & 16.8  & 77.47 & 69.71 & 11.08 & 23.82 & 77.97 & 67.26 \\
		GSWLCM  \cite{qiu2022global} & GRSL'20 & $\textemdash$ & $\textemdash$ & 12.82 & 13.72 & 70.24 & 13.92 & 21.62 & 19.85 & 69.41 & 42.71 & 10.82 & 18.71 & 67.32 & 53.31 \\
		\hline
		\multicolumn{16}{l}{\textit{Low-rank and Sparse Decomposition}}  \\
		\hline
		IPI \cite{gao2013infrared}   & TIP'13    & $\textemdash$ & $\textemdash$ & 27.92 & 30.12 & 81.37 & 16.18 & 25.16 & 34.64 & 76.26 & 43.41 & 28.63 & 38.18 & 74.49 & 41.23 \\
		NIPPS \cite{dai2017non}  & IPT'17    & $\textemdash$ & $\textemdash$ & 5.424 & 8.52  & 65.47 & 31.59 & 17.03 & 27.54 & 76.88 & 52.01 & 30.21 & 40.94 & 89.41 & 35.18 \\
		RIPT \cite{dai2017reweighted}  & JSTARS'17 & $\textemdash$ & $\textemdash$ & 14.11 & 17.43 & 77.55 & 28.31 & 24.13 & 33.98 & 78.54 & 56.24 & 29.17 & 36.12 & 91.85 & 344.3 \\
		NRAM \cite{zhang2018infrared}  & RS'18     & $\textemdash$ & $\textemdash$ & 9.882 & 18.71 & 72.48 & 24.73 & 8.972 & 15.27 & 71.47 & 68.98 & 12.08 & 18.61 & 72.58 & 84.77 \\
		NOLC \cite{zhang2019infrared1}  & RS'19     & $\textemdash$ & $\textemdash$ & 12.39 & 22.18 & 75.38 & 21.94 & 12.67 & 20.87 & 74.66 & 67.31 & 23.87 & 34.9  & 85.47 & 58.2  \\
		PSTNN \cite{zhang2019infrared} & RS'19     & $\textemdash$ & $\textemdash$ & 24.57 & 28.71 & 71.99 & 35.26 & 19.14 & 27.16 & 73.14 & 61.58 & 27.72 & 39.8  & 66.13 & 44.17 \\
		\hline
		\multicolumn{16}{l}{\textit{Deep Learning Methods}}  \\
		\hline
		MDvsFA \cite{wang2019miss}                  & ICCV'19  & 3.919M                         & 998.6G  & 49.50                     & 47.41  & 82.11                     & 80.33 &  $\textemdash$ &  $\textemdash$ &  $\textemdash$ &  $\textemdash$ & 75.14 &  $\textemdash$ & 90.47 & 25.34 \\
		ACM  \cite{Dai_2021_WACV}                     & WACV'21 & 0.5198M                        & 2.009G  & 63.39                     & 60.81  & 91.25                     & 8.961 & 73.84  & 69.83  & 97.52  & 76.35  & 68.48                     & 69.26  & 96.26  & 10.27  \\
		ALCNet  \cite{ALCNet}                  & TGRS'21 & 0.5404M                        & 2.127G  & 62.05                     & 59.58  & 92.19                     & 31.56 &  $\textemdash$ &  $\textemdash$ &  $\textemdash$ &  $\textemdash$ & 61.13   & 60.61  & 96.51 & 29.09 \\
		FC3-Net  \cite{FC3-Net}                 & MM'22   & 6.896M                         & 10.57G  & 64.98                     & 63.59  & 92.93                     & 15.73 & $\textemdash$ & $\textemdash$ & $\textemdash$ & $\textemdash$ & $\textemdash$                    & $\textemdash$ & $\textemdash$ & $\textemdash$ \\
		DNANet \cite{DNANet}                   & TIP'22  & 4.6968M                        & 56.08G  & 68.87                     & 67.53  & 94.95                     & 13.38 &  \textbf{\textcolor{green}{74.88}}  & 70.23  & 97.8   & 30.07  & 92.67                     & 92.09  & \textbf{\textcolor{green}{99.53}}  & 2.347  \\
		ISNet \cite{ISNet}                    & CVPR'22 & 1.09M                          & 121.90G & 68.77                     & 64.84  & 95.56                     & 15.39 & 72.50   & 70.84  & 98.41  & 28.61  & 89.81                     & 88.92  & 99.12  & 4.211  \\
		AGPCNet \cite{AGPCNet}                  & TAES'23 & 12.36M                         & 327.54G & 68.81                     & 66.18  & 94.26                     & 15.85 & 74.71  & \textbf{\textcolor{green}{71.49}}  & 97.67  & 34.84  & 88.71                     & 87.48  & 97.57  & 7.541  \\
		UIUNet  \cite{UIUNet}                  & TIP'23  & 50.54M                         & 217.85G & 69.13                     & 67.19  & 94.27                     & 16.47 & 74.24  & 70.57  & 98.35  & 23.13  & 90.77                     & 90.17  & 99.29  & 2.39   \\
		RDIAN  \cite{RDIAN}                   & TGRS'23 & 0.131M                         & 14.76G  & 64.37                     & 64.90   & 92.26                     & 18.2  & 74.19  & 69.8   & \textbf{\textcolor{green}{99.17}}  & 23.97  & 81.06                     & 81.72  & 98.36  & 14.09  \\
		MTUNet \cite{MTUNet}                   & TGRS'23 & 4.07M                          & 24.43G  & 67.50                      & 66.15  & 93.27                     & 14.75 & 74.70   & 70.66  & 98.49  & 39.73  & 87.49                     & 87.70   & 98.60   & 3.76   \\
		ABC   \cite{ABC}                    & ICME'23 & 73.51M                       & 332.55G & 72.02                     & 68.81  & 93.76                     & 9.457 &\textbf{\textcolor{cyan}{76.12}}   & \textbf{\textcolor{orange}{71.83}}  & \textbf{\textcolor{cyan}{99.59}}  & 20.33  & 92.85                     & 92.45  & 99.29  & 2.90    \\
		SRNet  \cite{SRNet}                   & TMM'23  & 0.4045M & $\textemdash$  & 69.45                     & 65.51  & 96.77                    & 13.05 & $\textemdash$ & $\textemdash$ & $\textemdash$ & $\textemdash$ & $\textemdash$                    & $\textemdash$ & $\textemdash$ & $\textemdash$ \\
		SeRankDet  \cite{SeRankDet}               & TGRS'24 & 108.89M                        & 568.74G & \textbf{\textcolor{orange}{73.66}}                     & \textbf{\textcolor{orange}{69.11}}  & 94.28                     & 8.37  & \textbf{\textcolor{orange}{75.49}}  & 71.08  & 98.97  & 18.9   & \textbf{\textcolor{green}{94.28}}                     & 93.69  & \textbf{\textcolor{cyan}{99.77}}  & \textbf{\textcolor{green}{2.03}}   \\
		MSHNet  \cite{MSHNet}                  & CVPR'24 & 4.07M                          & 24.43G  & 67.87 & 61.70   & 92.86 & 8.88  & $\textemdash$ & $\textemdash$ & $\textemdash$ & $\textemdash$ & 80.55 & $\textemdash$ & 97.99  & 11.77  \\
		$\text{Mrf}^3\text{Net}$ \cite{MRF3Net} & TGRS'24 & 0.538M                         & 33.2G   & 67.79                     & 68.74  & 95.28                     & 14.5  & $\textemdash$ & $\textemdash$ & $\textemdash$ & $\textemdash$ & \textbf{\textcolor{orange}{95.21}}                     & \textbf{\textcolor{orange}{95.23}}  & 99.36  & \textbf{\textcolor{orange}{1.86}}   \\
		SCTransNet \cite{SCTransNet}               & TGRS'24 & 11.19M                         & 67.4G   & 68.03                     & 68.15  & 93.27                     & 10.74 & $\textemdash$ & $\textemdash$ & $\textemdash$ & $\textemdash$ & 94.09                     & \textbf{\textcolor{green}{94.38}}  & 98.62  & 4.29   \\
		CSRNet \cite{CSRNet}       & TIP'24  & 0.4045M                        & 121G    & 65.87                     & 66.70   & \textbf{\textcolor{cyan}{98.16}}                     & 12.08 & $\textemdash$ & $\textemdash$ & $\textemdash$ & $\textemdash$ & $\textemdash$                    & $\textemdash$ & $\textemdash$ & $\textemdash$ \\
		\hline
		\multicolumn{16}{l}{\textit{Deep Unfolding-Based Methods}}  \\
		\hline
		RPCANet  \cite{RPCANet}                 & WACV'24 & 0.68M                          & 179.74G  & 63.21                     & 64.27 & 88.31                     & \textbf{\textcolor{orange}{4.39}}  & 72.54  & $\textemdash$ & 98.21  & 34.14  & 89.31                     & 89.03 & 97.14  & 2.87   \\
		\hline
		\multicolumn{16}{l}{\textit{Fine-tuning Models Based on Segment Anything Model} \cite{SAM}}   \\
		\hline
		IRSAM  \cite{IRSAM}                   & ECCV'24 & 12.33M                         & $\textemdash$  & \textbf{\textcolor{cyan}{73.69}}                     & \textbf{\textcolor{green}{68.97}}  & \textbf{\textcolor{green}{96.92}}                     & \textbf{\textcolor{green}{7.55}}  & $\textemdash$ & $\textemdash$ & $\textemdash$ & $\textemdash$ & 92.59                     & 93.29  & 98.87  & 6.94  \\
		\noalign{\hrule height 1pt} \rowcolor[rgb]{0.9,0.9,0.9}
		\multicolumn{16}{l}{\textit{Our proposed method}}  \\
		\hline \rowcolor[rgb]{0.9,0.9,0.9}
		LCRNet (Ours) & $\textemdash$ & 1.65M & 59.3G & \textbf{\textcolor{green}{72.67}} & \textbf{\textcolor{cyan}{70.19}} & \textbf{\textcolor{orange}{97.30}} & \textbf{\textcolor{cyan}{1.8}} & \textbf{\textcolor{orange}{75.49}} & \textbf{\textcolor{cyan}{71.93}} & \textbf{\textcolor{orange}{99.24}} & \textbf{\textcolor{cyan}{9.80}} & \textbf{\textcolor{cyan}{95.76}} & \textbf{\textcolor{cyan}{96.43}} & \textbf{\textcolor{orange}{99.55}} & \textbf{\textcolor{cyan}{1.05}} \\
		\noalign{\hrule height 1pt}
	\end{tabular}
\end{table*}
We compare the proposed method with related model-driven and data-driven methods. The model-driven methods include Background Suppression Methods and Low-rank and Sparse Decomposition, while the data-driven methods encompass Deep Learning Methods, Deep Unfolding-Based Methods, and Fine-tuning Models Based on the Segment Anything Model \cite{SAM}. Specifically, the compared algorithms are as follows:
\begin{itemize}
	\item Background Suppression Methods: NWMTH \cite{bai2010analysis}, ILCM \cite{GRSL2014ILCM}, MPCM \cite{PR16MPCM}, WLDM \cite{deng2016small}, RLCM \cite{GRSL2018RLCM}, FKRW \cite{qin2019infrared}, TLLCM \cite{han2019local}, GSWLCM  \cite{qiu2022global}.
	\item Low-rank and Sparse Decomposition: IPI \cite{gao2013infrared}, NIPPS \cite{dai2017non}, RIPT \cite{dai2017reweighted}, NRAM \cite{zhang2018infrared}, NOLC \cite{zhang2019infrared1}, PSTNN \cite{zhang2019infrared}.
	\item Deep Learning Methods: MDvsFA \cite{wang2019miss}, ACM  \cite{Dai_2021_WACV}, ALCNet  \cite{ALCNet}, FC3-Net  \cite{FC3-Net}, DNANet \cite{DNANet},  ISNet \cite{ISNet}, AGPCNet \cite{AGPCNet}, UIUNet  \cite{UIUNet}, RDIAN  \cite{RDIAN}, MTUNet \cite{MTUNet}, ABC   \cite{ABC}, SRNet  \cite{SRNet}, SeRankDet  \cite{SeRankDet}, MSHNet  \cite{MSHNet}, $\text{Mrf}^3\text{Net}$ \cite{MRF3Net}, SCTransNet \cite{SCTransNet}, CSRNet \cite{CSRNet}.
	\item Deep Unfolding-Based Methods: RPCANet \cite{RPCANet}.
	\item Fine-tuning Models Based on Segment Anything Model \cite{SAM}: IRSAM \cite{IRSAM}.
\end{itemize}

\subsubsection{Quantitative Evaluation} The results of different methods on three public datasets \cite{ISNet,AGPCNet,DNANet}  are listed in Table \ref{tab:sota}. From this table, we can observe that data-driven methods are significantly superior to model-driven methods. Meanwhile, our proposed LCRNet demonstrates superior performance compared to the other methods across three benchmarks, consistently ranking among the top three in terms of IoU, nIoU, $P_d$, and $F_a$ metrics. In particular, it achieves the best results in terms of nIoU and $F_a$, highlighting its advantages for infrared small target detection tasks.

The IRSTD-1k dataset exhibits significant multi-scale characteristics, with 18\% of targets in the range (0, 10], 46\% in (10, 40], 27\% in (40, 100], and 9\% in (100, $\infty$), making the task particularly challenging. On this dataset, our method outperforms 33 state-of-the-art approaches, achieving the best nIoU and $F_a$ scores. Although the target-level $P_d$ metric is slightly lower than CSRNet \cite{CSRNet} by 0.86\%, our method ranks second, while significantly outperforming CSRNet in $F_a$ and achieving 6.8\% higher pixel-level IoU and 3.49\% higher nIoU. Pixel-level IoU is only surpassed by IRSAM \cite{IRSAM} and SeRankDet \cite{SeRankDet}. The performance of IRSAM is due to its SAM fine-tuning, which favors large targets and boosts IoU. However, our method excels in nIoU, where IRSAM lags. SeRankDet's high performance is largely attributed to its higher parameter and computational cost (65.9 times more parameters and 9.6 times more computations). In summary, our method achieves a favorable balance between performance and resource efficiency, offering superior results with significantly lower computational cost.

On the SIRSTAUG dataset, where 97\% of targets occupy pixel areas in the range (0, 10], classifying them as typical small targets, our model again achieves the best performance in both nIoU and $F_a$. Although IoU is only slightly lower than ABC \cite{ABC} by 0.63\% and $P_d$ is 0.35\% lower, our method outperforms ABC in both nIoU and $F_a$. Notably, ABC has 44.5 times more parameters and 5.6 times higher computational cost compared to our method. This highlights the efficient trade-off between performance and resource usage in our approach. Furthermore, the results on the SIRSTAUG dataset demonstrate that LCRNet adapts well even when the target sizes are highly concentrated and small.

\begin{figure}
	\centering
	\includegraphics[width=\linewidth]{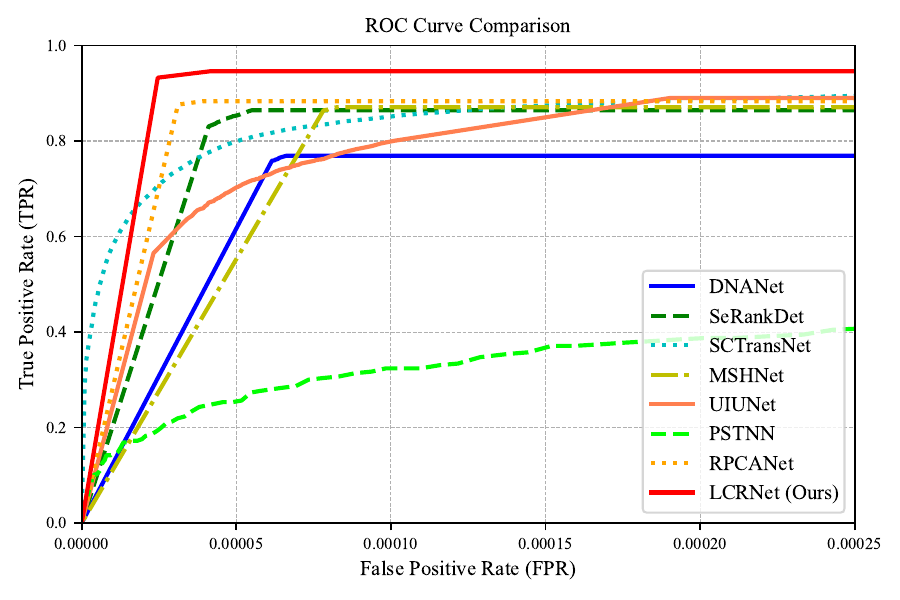}
	\caption{ROC curves of our LCRNet and other approaches on IRSTD-1k.}
	\label{fig:ROC}   
\end{figure}

On the NUDT-SIRST dataset, 25\% of the targets fall within (0, 10], 44\% within (10, 40], and 30\% within (40, 100], exhibiting cross-scale characteristics without any particularly large targets. On this dataset, the proposed LCRNet outperforms all compared approaches, achieving the best results in IoU, nIoU, and $F_a$ metrics, and second-best in $P_d$, only 0.22\% lower than the best performing method. It is evident that our method exhibits strong stability and robustness. To further verify this, we present the ROC curves in Fig. \ref{fig:ROC}, evaluated on the IRSTD-1k dataset, where our approach achieves the best AUC score. Additionally, due to the strong approximation ability of the adaptive multigrid method, our model is highly sensitive in capturing small targets, which may lead to an imbalance between miss detection and false alarm rates.

\subsubsection{Qualitative Evaluation}
\begin{figure}
	\centering
	\includegraphics[width=\linewidth]{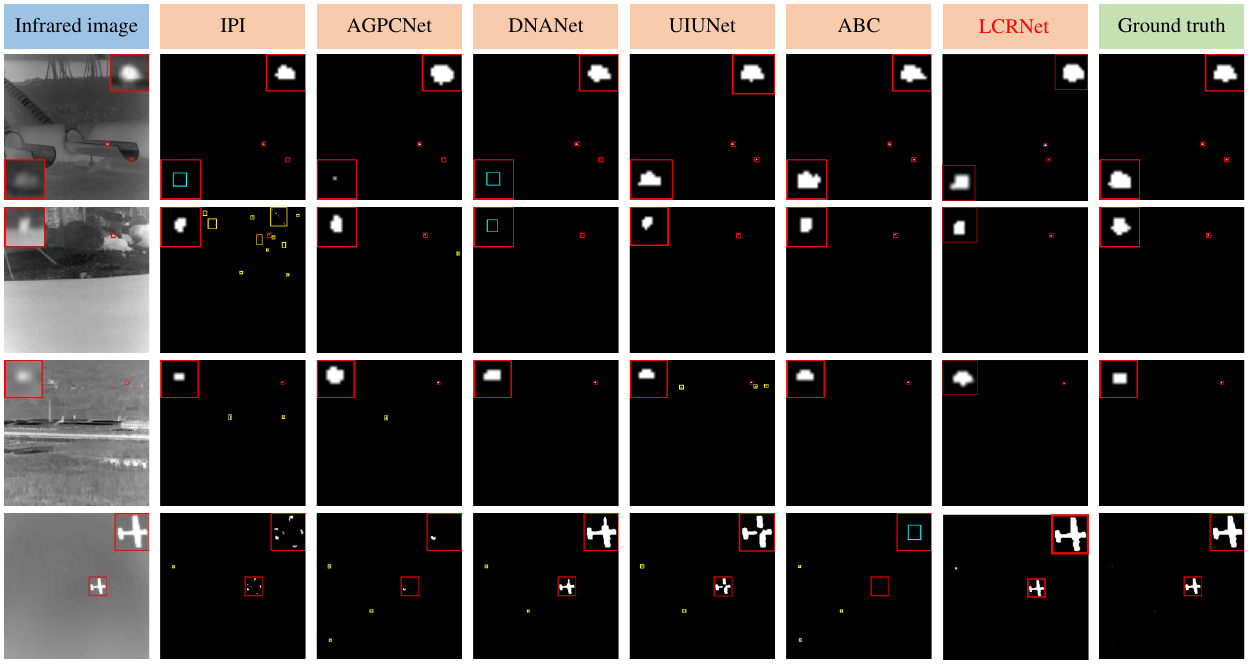}
	\caption{Visualization comparison of detection results via different methods on representative images from IRSTD-1k dataset. The red, yellow, and cyan boxes denote correct detections, false alarms, and missed detections, respectively.}
	\label{fig:visual}   
\end{figure}
The qualitative results presented in Fig. \ref{fig:visual} offer a visual comparison of the detection performance of various methods on representative images from the IRSTD-1k dataset. Traditional methods generate significantly more false alarms (highlighted by yellow boxes) and missed detections (highlighted by cyan boxes) compared to deep learning-based approaches, underscoring the superior performance of deep learning techniques in infrared small target detection. Deep neural networks are capable of learning rich, hierarchical features from the data, enabling them to effectively distinguish between true targets and background clutter. Among the deep learning-based methods, our proposed approach consistently outperforms other state-of-the-art methods, demonstrating its effectiveness in addressing the challenges of small target detection in infrared imagery. Moreover, our method adapts well to targets of varying sizes and signal-to-noise ratios, achieving excellent segmentation performance and low false alarm rates for both small and large targets.

\subsection{Ablation Study}
\begin{table*}
	\caption{Ablation for LCRNet on the IRSTD-1k dataset.}
	\centering
	\begin{tabular}{l|l|c c c c c c}
		\noalign{\hrule height 1pt}
		\multirow{2}{*}{Ablation} & \multirow{2}{*}{Variant} & \multicolumn{4}{c}{Evaluation Metrics}  \\
		\multicolumn{1}{l|}{} & \multicolumn{1}{l|}{}
		& IoU(\%)  $\uparrow$     & nIoU(\%) $\uparrow$   &  Pd(\%) $\uparrow$     & FA($10^{-6}$) $\downarrow$ \\
		\noalign{\hrule height 1pt}
		-- & LCRNet & 72.67 & 70.19 & 97.30 & 1.8 \\
		\hline
		\multirow{6}{*}{\tabincell{l}{C2FBlock}}  & C2FBlock $\rightarrow$ C2FBlock w/ FFN \cite{xie2021segformer} \tiny\textcolor{red}{(NIPS'21)}    & 70.49 \textcolor{gray}{(-2.18)} & 67.76 \textcolor{gray}{(-2.43)} & 93.94 \textcolor{gray}{(-3.36)} & 3.0 \textcolor{gray}{(+1.2)} \\
		& C2FBlock $\rightarrow$  C2FBlock w/ ConvFFN \cite{wang2021pyramid} \tiny\textcolor{red}{(ICCV'21)}    & 69.43 \textcolor{gray}{(-3.24)} & 67.75 \textcolor{gray}{(-2.44)} & 92.93 \textcolor{gray}{(-4.37)} & 2.6 \textcolor{gray}{(+0.8)} \\
		& C2FBlock w/ pre-BasicBlock \cite{huang2023revisiting} $\rightarrow$ BasicBlock \cite{he2016deep}    & 66.74 \textcolor{gray}{(-5.93)} & 66.97 \textcolor{gray}{(-3.22)} & 92.26 \textcolor{gray}{(-5.04)} & 15.8 \textcolor{gray}{(+14)} \\
		  & $L_i = \left\{3,3,3,3\right\}$ $\rightarrow$  $L_i = \left\{1,1,1,1\right\}$       & 71.70 \textcolor{gray}{(-0.97)} & 69.72 \textcolor{gray}{(-0.47)} & 95.29 \textcolor{gray}{(-2.01)} & 1.8 \textcolor{gray}{(+0.0)}  \\
		& $L_i = \left\{3,3,3,3\right\}$ $\rightarrow$  $L_i = \left\{2,2,2,2\right\}$    & 72.67 \textcolor{gray}{(-0.00)} & 70.03 \textcolor{gray}{(-0.16)} & 95.62 \textcolor{gray}{(-1.68)} & 2.5 \textcolor{gray}{(+0.7)} \\
		& $L_i = \left\{3,3,3,3\right\}$ $\rightarrow$  $L_i = \left\{4,4,4,4\right\}$    & 72.26 \textcolor{gray}{(-0.41)} & 69.16 \textcolor{gray}{(-1.03)} & 95.96 \textcolor{gray}{(-1.34)} & 1.9 \textcolor{gray}{(+0.1)} \\
		\hline
		\multirow{9}{*}{\tabincell{l}{Attention variants}} & DLC-Attention $\rightarrow$ LSKA \cite{li2023large,li2024lsknet} \tiny\textcolor{red}{(ICCV'23 \& IJCV'24)}    & 68.72 \textcolor{gray}{(-3.95)} & 68.37 \textcolor{gray}{(-1.82)} & 93.60 \textcolor{gray}{(-3.70)} & 4.2 \textcolor{gray}{(+2.4)} \\
		& DLC-Attention $\rightarrow$ LKA \cite{guo2023visual} \tiny\textcolor{red}{(CVM'23)}    & 67.35 \textcolor{gray}{(-5.32)} & 67.79 \textcolor{gray}{(-2.40)} & 92.59 \textcolor{gray}{(-4.71)} & 2.2 \textcolor{gray}{(+0.4)} \\
		& DLC-Attention $\rightarrow$  ConvMod \cite{hou2024conv2former} \tiny\textcolor{red}{(TPAMI'24)}    & 69.73 \textcolor{gray}{(-2.94)} & 68.75 \textcolor{gray}{(-1.44)} & 94.61 \textcolor{gray}{(-2.69)} & 2.2 \textcolor{gray}{(+0.4)}\\
		& DLC-Attention $\rightarrow$ SAA \cite{li2023moganet} \tiny\textcolor{red}{(ICLR'23)}   & 67.21 \textcolor{gray}{(-5.46)} & 68.33 \textcolor{gray}{(-1.86)} & 93.94 \textcolor{gray}{(-3.36)} & 1.9 \textcolor{gray}{(+0.1)} \\
		& DLC-Attention $\rightarrow$ Identity    & 66.90 \textcolor{gray}{(-5.77)} & 68.07 \textcolor{gray}{(-2.12)} & 94.28 \textcolor{gray}{(-3.02)} & 3.5 \textcolor{gray}{(+1.7)} \\
		& DLC-Attention w/o SiLU    & 70.78 \textcolor{gray}{(-1.98)} & 69.19 \textcolor{gray}{(-1.00)} & 95.29 \textcolor{gray}{(-2.01)} & 1.8 \textcolor{gray}{(+0.0)} \\
		& DLC-Attention w/o learnable temperature parameter    & 71.06 \textcolor{gray}{(-1.61)} & 69.49 \textcolor{gray}{(-0.70)} & 95.29 \textcolor{gray}{(-2.01)} & 3.2 \textcolor{gray}{(+1.4)} \\
		& DLC-Attention w/o split-attention     & 70.89 \textcolor{gray}{(-1.78)} & 68.33 \textcolor{gray}{(-1.86)} & 92.93 \textcolor{gray}{(-4.37)} & 1.4 \textcolor{gray}{(-0.4)}\\
		& DLC-Attention w/o residual connection   & 71.00 \textcolor{gray}{(-1.67)} & 69.03 \textcolor{gray}{(-1.16)} & 94.28 \textcolor{gray}{(-3.02)} & 4.4 \textcolor{gray}{(+2.6)} \\
		\hline
		\multirow{4}{*}{\tabincell{l}{Large Kernel variants}}  & HLKConv $\rightarrow$  Large Kernel \cite{guo2023visual} \tiny\textcolor{red}{(CVM'23)}        & 70.55 \textcolor{gray}{(-2.12)} & 68.51 \textcolor{gray}{(-1.68)} & 93.27 \textcolor{gray}{(-4.03)} & 1.7 \textcolor{gray}{(-0.1)}  \\
		& HLKConv $\rightarrow$  Large Separable Kernel \cite{lau2024large} \tiny\textcolor{red}{(Expert Syst. Appl'24)}    & 69.42 \textcolor{gray}{(-3.25)} & 68.02 \textcolor{gray}{(-2.17)} & 93.94 \textcolor{gray}{(-3.36)} & 2.4 \textcolor{gray}{(+0.6)} \\
		& HLKConv $\rightarrow$ $\text{DWConv}_{7\times7}$ \cite{liu2022convnet} \tiny\textcolor{red}{(CVPR'22)}   & 70.56 \textcolor{gray}{(-2.11)} & 68.60 \textcolor{gray}{(-1.59)} & 94.95 \textcolor{gray}{(-2.35)} & 2.0 \textcolor{gray}{(+0.2)} \\
		& $\text{HLKConv}_{7\times7} : \left\{3,3,2\right\}$ $\rightarrow$ $\text{HLKConv}_{11\times11} : \left\{3,5,2\right\}$       & 70.94 \textcolor{gray}{(-1.73)} & 69.46 \textcolor{gray}{(-0.73)} & 95.29 \textcolor{gray}{(-2.01)} & 2.2 \textcolor{gray}{(+0.4)}  \\
		\hline
		\multirow{2}{*}{\tabincell{l}{Number of groups}}  & $K = 4$ $\rightarrow$ $K = 1$   & 70.12 \textcolor{gray}{(-2.55)} & 68.71 \textcolor{gray}{(-1.48)} & 94.28 \textcolor{gray}{(-3.02)} & 1.7 \textcolor{gray}{(-0.1)}  \\
		& $K = 4$ $\rightarrow$ $K = 2$   & 71.69 \textcolor{gray}{(-0.98)} & 68.35 \textcolor{gray}{(-1.84)} & 93.94 \textcolor{gray}{(-3.36)} & 2.3 \textcolor{gray}{(+0.5)}  \\
		\hline
		\multirow{2}{*}{\tabincell{l}{Branch output scaling}}  & LayerScale \cite{touvron2021going} $\rightarrow$ None  & 70.79 \textcolor{gray}{(-1.88)} & 69.01 \textcolor{gray}{(-1.13)} & 94.95 \textcolor{gray}{(-2.35)} & 1.5 \textcolor{gray}{(-0.3)} \\
		& LayerScale \cite{touvron2021going} $\rightarrow$ ResScale \cite{shleifer2021normformer} & 71.52 \textcolor{gray}{(-1.15)} & 68.99 \textcolor{gray}{(-1.20)} & 94.28 \textcolor{gray}{(-3.02)} & 1.8 \textcolor{gray}{(+0.0)} \\
		\noalign{\hrule height 1pt}
	\end{tabular}
	\label{tab:Ablation}
\end{table*}

\subsubsection{The Effect of C2FBlock} We first evaluate our C2FBlock and present the results in Tab. \ref{tab:Ablation}. Specifically, we replace the BasicBlock in C2FBlock with FFN \cite{xie2021segformer} and ConvFFN \cite{wang2021pyramid} to eliminate the influence of the multigrid iteration simulation, resulting in a conventional MetaFormer-style Block \cite{yu2022metaformer, yu2023metaformer}. The results show that our C2FBlock outperforms the alternatives in IoU, nIoU, \(P_d\), and \(F_a\). This improvement can be attributed to the ability of C2FBlock to simulate an iterative process from coarse to fine at the spatial level, which helps in separating high-frequency signals—crucial for infrared small targets, which can be considered as a special type of high-frequency signal. Furthermore, we replace the pre-BasicBlock with a conventional BasicBlock in an ablation experiment. The results show that using pre-BasicBlock leads to a 5.93\% increase in IoU, a 3.22\% increase in nIoU, a 5.04\% improvement in \(P_d\), and a significant reduction in \(F_a\). Importantly, both configurations have identical parameter counts, computational complexity, and memory usage, further validating the effectiveness of the C2FBlock incorporating the multigrid method concept.

We then perform hyperparameter tuning for the C2FBlock. From Tab. \ref{tab:Ablation}, it is evident that $L_i = \{3,3,3,3\}$ achieves the best performance. However, it is noteworthy that when $L_i = \{1,1,1,1\}$, the model still demonstrates excellent performance, with a parameter count of 854.67K and a computational cost of 32.14G.

\subsubsection{The Effect of DLC-Attention} DLC-Attention is a key component of the proposed model. To validate its effectiveness, we conduct comparisons in two aspects: (1) comparing it with similar large-kernel attention mechanisms \cite{guo2023visual,li2023large,li2024lsknet,hou2024conv2former,li2023moganet}, and (2) performing ablation studies on its internal components.

First, we compare our proposed DLC-Attention with similar large-kernel attention mechanisms. As shown in the results in Tab. \ref{tab:Ablation}, the proposed DLC-Attention outperforms existing attention mechanisms in several key metrics. Compared to LKA \cite{guo2023visual}, our method improves by 5.32\%@IoU, 2.40\%@nIoU, and 4.71\%@$P_d$, as it better distinguishes targets from the background within long-range dependencies while preserving fine-grained details in small regions. This leads to improved target-level and pixel-level performance, along with adaptive receptive field allocation based on target shape.
When compared to LSKA \cite{li2023large,li2024lsknet}, our method shows improvements of 3.95\%@IoU, 1.82\%@nIoU, and 3.70\%@$P_d$. Additionally, it surpasses ConvMod \cite{hou2024conv2former} by 2.94\%@IoU, 1.44\%@nIoU, and 2.69\%@$P_d$, thanks to its broader receptive field range and reduced channel redundancy through higher-order interactions.
Finally, against SAA \cite{li2023moganet}, our approach achieves gains of 5.46\%@IoU, 1.86\%@nIoU, and 3.36\%@$P_d$. This is largely due to HLKConv, which mitigates block-like artifacts from dilated convolutions, reducing their negative impact on small target detection.

Next, we perform an ablation study on the internal components of DLC-Attention. When DLC-Attention is replaced with the Identity function, a significant drop in performance is observed, confirming the effectiveness of the proposed attention mechanism. Additionally, the ablation of any individual component results in a decrease in performance. Among these, the absence of the nonlinear activation function has the most substantial impact, which can be attributed to the shallow nature of the model. The nonlinear activation function provides greater nonlinearity and abstraction, which is crucial for capturing complex patterns in the data. Furthermore, the residual connection plays a critical role in mitigating the overconfidence issues of the SoftMax function, reducing its negative impact on model performance.

\subsubsection{The Effect of HLKConv} 
To validate the effectiveness of HLKConv, we compare it with three similar large-kernel convolutions and conduct hyperparameter tuning on kernel sizes. The results are summarized in Tab. \ref{tab:Ablation}.
Our HLKConv achieves performance improvements of 2.12\%@IoU, 1.68\%@nIoU, and 4.03\%@$P_d$ compared to LKConv \cite{guo2023visual}, which only uses the large-kernel decomposition rule. This improvement is attributed to the hierarchical convolution approach that mitigates the negative impact of dilated convolutions. Compared to LSKConv \cite{lau2024large}, HLKConv improves by 3.25\%@IoU, 2.17\%@nIoU, and 3.36\%@$P_d$, as LSKConv's low-rank assumption on the convolution parameter space often leads to performance degradation when it does not align with the actual data. When compared to $\text{DWConv}_{7\times7}$ \cite{liu2022convnet}, HLKConv shows improvements of 2.11\%@IoU, 1.59\%@nIoU, and 2.35\%@$P_d$, as DWConv's sparsity is insufficient for effective training optimization.
Additionally, when the kernel size is increased from 7 to 11, a performance drop is observed. This is due to the fact that when the kernel size exceeds the feature map dimensions, the convolution operation itself becomes global, reducing the inductive bias, which necessitates more data or more complex data augmentation strategies \cite{ding2022scaling}.

\subsubsection{Model Size Discussion} This section discusses the impact of the number of groups and branch output scaling. The results in Tab. \ref{tab:Ablation} show that increasing the number of groups improves overall performance. However, excessively large group numbers can lead to a decrease in inference speed. Additionally, the effectiveness of the LayerScale mechanism  \cite{touvron2021going} in enhancing performance is also validated.

\subsection{Limitations}
Although LCRNet incorporates many DWConv operations, reducing the model's overall computational cost, it introduces a low computation-to-memory access ratio \cite{ma2018shufflenet}, which can decrease speed on modern architectures. However, due to the model's relatively small width, low number of groups, and use of smaller kernels for large convolutions, the speed reduction remains within an acceptable range. Consistent with previous infrared small target detection inference speed testing methods \cite{DNANet,CSRNet}, we achieve an inference speed of 20.3 FPS on a NVIDIA GeForce RTX 3080ti GPU using PyTorch, with a batch size of 8 and an input size of 512 × 512. Note that the current DWConv implementation in PyTorch is not fully optimized \cite{ding2024unireplknet}. According to the roofline model \cite{ma2018shufflenet}, as kernel size increases, computational density also increases, meaning latency will not rise as sharply as FLOPs. For real-time deployment, optimized CUDA implementations \cite{ding2022scaling} can be considered.

\section{Conclusion}\label{Section:Conclusion}
In this work, we propose a novel method for ISTD by leveraging two key priors of infrared images. Our approach introduces three key components: C2FBlock, DLC-Attention and HLKConv, which are extensively evaluated across multiple publicly available datasets. Experimental results demonstrate that our method outperforms existing SOTA approaches on several performance metrics.
Notably, our method is simple and elegant, avoiding complex strategies like deep supervision, yet it achieves significant improvements. This highlights the untapped potential of the U-Net architecture for ISTD, suggesting further opportunities for refinement within this framework.
Future research could explore specialized downsampling and upsampling techniques, dynamic weighting strategies for deep supervision, and novel architectural designs to further enhance performance in challenging infrared environments.

\bibliographystyle{IEEEtran}
\bibliography{reference}

\begin{thebibliography}{10}
\providecommand{\url}[1]{#1}
\csname url@samestyle\endcsname
\providecommand{\newblock}{\relax}
\providecommand{\bibinfo}[2]{#2}
\providecommand{\BIBentrySTDinterwordspacing}{\spaceskip=0pt\relax}
\providecommand{\BIBentryALTinterwordstretchfactor}{4}
\providecommand{\BIBentryALTinterwordspacing}{\spaceskip=\fontdimen2\font plus
\BIBentryALTinterwordstretchfactor\fontdimen3\font minus
  \fontdimen4\font\relax}
\providecommand{\BIBforeignlanguage}[2]{{%
\expandafter\ifx\csname l@#1\endcsname\relax
\typeout{** WARNING: IEEEtran.bst: No hyphenation pattern has been}%
\typeout{** loaded for the language `#1'. Using the pattern for}%
\typeout{** the default language instead.}%
\else
\language=\csname l@#1\endcsname
\fi
#2}}
\providecommand{\BIBdecl}{\relax}
\BIBdecl

\bibitem{FEAIRSTD}
S.~Chen, H.~Wang, Z.~Shen, G.~Zhang, C.~Ning, and X.~Zhang, ``A feature
  enhancement and augmentation-based infrared small target detection network,''
  \emph{IEEE Geoscience and Remote Sensing Letters}, vol.~21, pp. 1--5, 2024.

\bibitem{zhao2022single}
M.~Zhao, W.~Li, L.~Li, J.~Hu, P.~Ma, and R.~Tao, ``Single-frame infrared
  small-target detection: A survey,'' \emph{IEEE Geoscience and Remote Sensing
  Magazine}, vol.~10, no.~2, pp. 87--119, 2022.

\bibitem{strickland2023infrared}
R.~N. Strickland, ``Infrared techniques for military applications,'' in
  \emph{Infrared Methodology and Technology}.\hskip 1em plus 0.5em minus
  0.4em\relax CRC Press, 2023, pp. 397--427.

\bibitem{yi2023spatial}
H.~Yi, C.~Yang, R.~Qie, J.~Liao, F.~Wu, T.~Pu, and Z.~Peng, ``Spatial-temporal
  tensor ring norm regularization for infrared small target detection,''
  \emph{IEEE Geoscience and Remote Sensing Letters}, vol.~20, pp. 1--5, 2023.

\bibitem{SeRankDet}
Y.~Dai, P.~Pan, Y.~Qian, Y.~Li, X.~Li, J.~Yang, and H.~Wang, ``Pick of the
  bunch: Detecting infrared small targets beyond hit-miss trade-offs via
  selective rank-aware attention,'' \emph{IEEE Transactions on Geoscience and
  Remote Sensing}, vol.~62, pp. 1--15, 2024.

\bibitem{liu2023infrared}
F.~Liu, C.~Gao, F.~Chen, D.~Meng, W.~Zuo, and X.~Gao, ``Infrared small and dim
  target detection with transformer under complex backgrounds,'' \emph{IEEE
  Transactions on Image Processing}, vol.~32, pp. 5921--5932, 2023.

\bibitem{ISNet}
M.~Zhang, R.~Zhang, Y.~Yang, H.~Bai, J.~Zhang, and J.~Guo, ``Isnet: Shape
  matters for infrared small target detection,'' in \emph{2022 IEEE/CVF
  Conference on Computer Vision and Pattern Recognition (CVPR)}, 2022, pp.
  867--876.

\bibitem{SRNet}
F.~Lin, S.~Ge, K.~Bao, C.~Yan, and D.~Zeng, ``Learning shape-biased
  representations for infrared small target detection,'' \emph{IEEE
  Transactions on Multimedia}, vol.~26, pp. 4681--4692, 2024.

\bibitem{CSRNet}
F.~Lin, K.~Bao, Y.~Li, D.~Zeng, and S.~Ge, ``Learning contrast-enhanced
  shape-biased representations for infrared small target detection,''
  \emph{IEEE Transactions on Image Processing}, vol.~33, pp. 3047--3058, 2024.

\bibitem{chen2022local}
F.~Chen, C.~Gao, F.~Liu, Y.~Zhao, Y.~Zhou, D.~Meng, and W.~Zuo, ``Local patch
  network with global attention for infrared small target detection,''
  \emph{IEEE Transactions on Aerospace and Electronic Systems}, vol.~58, no.~5,
  pp. 3979--3991, 2022.

\bibitem{MSHNet}
Q.~Liu, R.~Liu, B.~Zheng, H.~Wang, and Y.~Fu, ``Infrared small target detection
  with scale and location sensitivity,'' in \emph{Proceedings of the IEEE/CVF
  Computer Vision and Pattern Recognition}, 2024.

\bibitem{DNANet}
B.~Li, C.~Xiao, L.~Wang, Y.~Wang, Z.~Lin, M.~Li, W.~An, and Y.~Guo, ``Dense
  nested attention network for infrared small target detection,'' \emph{IEEE
  Transactions on Image Processing}, vol.~32, pp. 1745--1758, 2023.

\bibitem{UIUNet}
X.~Wu, D.~Hong, and J.~Chanussot, ``Uiu-net: U-net in u-net for infrared small
  object detection,'' \emph{IEEE Transactions on Image Processing}, vol.~32,
  pp. 364--376, 2023.

\bibitem{ALCNet}
Y.~Dai, Y.~Wu, F.~Zhou, and K.~Barnard, ``Attentional local contrast networks
  for infrared small target detection,'' \emph{IEEE Transactions on Geoscience
  and Remote Sensing}, vol.~59, no.~11, pp. 9813--9824, 2021.

\bibitem{Dai_2021_WACV}
------, ``Asymmetric contextual modulation for infrared small target
  detection,'' in \emph{Proceedings of the IEEE/CVF Winter Conference on
  Applications of Computer Vision (WACV)}, January 2021, pp. 950--959.

\bibitem{RDIAN}
H.~Sun, J.~Bai, F.~Yang, and X.~Bai, ``Receptive-field and direction induced
  attention network for infrared dim small target detection with a large-scale
  dataset irdst,'' \emph{IEEE Transactions on Geoscience and Remote Sensing},
  vol.~61, pp. 1--13, 2023.

\bibitem{MTUNet}
T.~Wu, B.~Li, Y.~Luo, Y.~Wang, C.~Xiao, T.~Liu, J.~Yang, W.~An, and Y.~Guo,
  ``Mtu-net: Multilevel transunet for space-based infrared tiny ship
  detection,'' \emph{IEEE Transactions on Geoscience and Remote Sensing},
  vol.~61, pp. 1--15, 2023.

\bibitem{ABC}
P.~Pan, H.~Wang, C.~Wang, and C.~Nie, ``Abc: Attention with bilinear
  correlation for infrared small target detection,'' in \emph{2023 IEEE
  International Conference on Multimedia and Expo (ICME)}, 2023, pp.
  2381--2386.

\bibitem{SCTransNet}
S.~Yuan, H.~Qin, X.~Yan, N.~Akhtar, and A.~Mian, ``Sctransnet: Spatial-channel
  cross transformer network for infrared small target detection,'' \emph{IEEE
  Transactions on Geoscience and Remote Sensing}, vol.~62, pp. 1--15, 2024.

\bibitem{IRSAM}
M.~Zhang, Y.~Wang, J.~Guo, Y.~Li, X.~Gao, and J.~Zhang, ``Irsam: Advancing
  segment anything model for infrared small target detection,'' in
  \emph{Computer Vision -- ECCV 2024}, A.~Leonardis, E.~Ricci, S.~Roth,
  O.~Russakovsky, T.~Sattler, and G.~Varol, Eds.\hskip 1em plus 0.5em minus
  0.4em\relax Cham: Springer Nature Switzerland, 2025, pp. 233--249.

\bibitem{SAM}
A.~Kirillov, E.~Mintun, N.~Ravi, H.~Mao, C.~Rolland, L.~Gustafson, T.~Xiao,
  S.~Whitehead, A.~C. Berg, W.-Y. Lo, P.~Dollár, and R.~Girshick, ``Segment
  anything,'' in \emph{2023 IEEE/CVF International Conference on Computer
  Vision (ICCV)}, 2023, pp. 3992--4003.

\bibitem{GRSL2014ILCM}
J.~Han, Y.~Ma, B.~Zhou, F.~Fan, K.~Liang, and Y.~Fang, ``A robust infrared
  small target detection algorithm based on human visual system,'' \emph{IEEE
  Geoscience and Remote Sensing Letters}, vol.~11, no.~12, pp. 2168--2172,
  2014.

\bibitem{gao2013infrared}
C.~Gao, D.~Meng, Y.~Yang, Y.~Wang, X.~Zhou, and A.~G. Hauptmann, ``Infrared
  patch-image model for small target detection in a single image,'' \emph{IEEE
  Transactions on Image Processing}, vol.~22, no.~12, pp. 4996--5009, 2013.

\bibitem{luo2016understanding}
W.~Luo, Y.~Li, R.~Urtasun, and R.~Zemel, ``Understanding the effective
  receptive field in deep convolutional neural networks,'' \emph{Advances in
  neural information processing systems}, vol.~29, 2016.

\bibitem{he2019ode}
X.~He, Z.~Mo, P.~Wang, Y.~Liu, M.~Yang, and J.~Cheng, ``Ode-inspired network
  design for single image super-resolution,'' in \emph{Proceedings of the
  IEEE/CVF Conference on Computer Vision and Pattern Recognition}, 2019, pp.
  1732--1741.

\bibitem{10061442}
Z.~Zhang, K.~Chen, K.~Tang, and Y.~Duan, ``Fast multi-grid methods for
  minimizing curvature energies,'' \emph{IEEE Transactions on Image
  Processing}, vol.~32, pp. 1716--1731, 2023.

\bibitem{guo2023visual}
M.-H. Guo, C.-Z. Lu, Z.-N. Liu, M.-M. Cheng, and S.-M. Hu, ``Visual attention
  network,'' \emph{Computational Visual Media}, vol.~9, no.~4, pp. 733--752,
  2023.

\bibitem{bai2010analysis}
X.~Bai and F.~Zhou, ``Analysis of new top-hat transformation and the
  application for infrared dim small target detection,'' \emph{Pattern
  Recognition}, vol.~43, no.~6, pp. 2145--2156, 2010.

\bibitem{PR16MPCM}
Y.~Wei, X.~You, and H.~Li, ``Multiscale patch-based contrast measure for small
  infrared target detection,'' \emph{Pattern Recognition}, vol.~58, pp.
  216--226, 2016.

\bibitem{deng2016small}
H.~Deng, X.~Sun, M.~Liu, C.~Ye, and X.~Zhou, ``Small infrared target detection
  based on weighted local difference measure,'' \emph{IEEE Transactions on
  Geoscience and Remote Sensing}, vol.~54, no.~7, pp. 4204--4214, 2016.

\bibitem{GRSL2018RLCM}
J.~Han, K.~Liang, B.~Zhou, X.~Zhu, J.~Zhao, and L.~Zhao, ``Infrared small
  target detection utilizing the multiscale relative local contrast measure,''
  \emph{IEEE Geoscience and Remote Sensing Letters}, vol.~15, no.~4, pp.
  612--616, 2018.

\bibitem{qin2019infrared}
Y.~Qin, L.~Bruzzone, C.~Gao, and B.~Li, ``Infrared small target detection based
  on facet kernel and random walker,'' \emph{IEEE Transactions on Geoscience
  and Remote Sensing}, vol.~57, no.~9, pp. 7104--7118, 2019.

\bibitem{han2019local}
J.~Han, S.~Liu, G.~Qin, Q.~Zhao, H.~Zhang, and N.~Li, ``A local contrast method
  combined with adaptive background estimation for infrared small target
  detection,'' \emph{IEEE Geoscience and Remote Sensing Letters}, vol.~16,
  no.~9, pp. 1442--1446, 2019.

\bibitem{qiu2022global}
Z.~Qiu, Y.~Ma, F.~Fan, J.~Huang, and L.~Wu, ``Global sparsity-weighted local
  contrast measure for infrared small target detection,'' \emph{IEEE Geoscience
  and Remote Sensing Letters}, vol.~19, pp. 1--5, 2022.

\bibitem{dai2017non}
Y.~Dai, Y.~Wu, Y.~Song, and J.~Guo, ``Non-negative infrared patch-image model:
  Robust target-background separation via partial sum minimization of singular
  values,'' \emph{Infrared Physics \& Technology}, vol.~81, pp. 182--194, 2017.

\bibitem{dai2017reweighted}
Y.~Dai and Y.~Wu, ``Reweighted infrared patch-tensor model with both nonlocal
  and local priors for single-frame small target detection,'' \emph{IEEE
  journal of selected topics in applied earth observations and remote sensing},
  vol.~10, no.~8, pp. 3752--3767, 2017.

\bibitem{zhang2018infrared}
L.~Zhang, L.~Peng, T.~Zhang, S.~Cao, and Z.~Peng, ``Infrared small target
  detection via non-convex rank approximation minimization joint $\ell_{2,1}$
  norm,'' \emph{Remote Sensing}, vol.~10, no.~11, p. 1821, 2018.

\bibitem{zhang2019infrared1}
T.~Zhang, H.~Wu, Y.~Liu, L.~Peng, C.~Yang, and Z.~Peng, ``Infrared small target
  detection based on non-convex optimization with $l_p$-norm constraint,''
  \emph{Remote Sensing}, vol.~11, no.~5, p. 559, 2019.

\bibitem{zhang2019infrared}
L.~Zhang and Z.~Peng, ``Infrared small target detection based on partial sum of
  the tensor nuclear norm,'' \emph{Remote Sensing}, vol.~11, no.~4, p. 382,
  2019.

\bibitem{AGPCNet}
T.~Zhang, L.~Li, S.~Cao, T.~Pu, and Z.~Peng, ``Attention-guided pyramid context
  networks for detecting infrared small target under complex background,''
  \emph{IEEE Transactions on Aerospace and Electronic Systems}, vol.~59, no.~4,
  pp. 4250--4261, 2023.

\bibitem{li2023ilnet}
H.~Li, J.~Yang, R.~Wang, and Y.~Xu, ``Ilnet: Low-level matters for salient
  infrared small target detection,'' \emph{arXiv preprint arXiv:2309.13646},
  2023.

\bibitem{huang2023large}
T.~Huang, L.~Yin, Z.~Zhang, L.~Shen, M.~Fang, M.~Pechenizkiy, Z.~Wang, and
  S.~Liu, ``Are large kernels better teachers than transformers for convnets?''
  in \emph{International Conference on Machine Learning}.\hskip 1em plus 0.5em
  minus 0.4em\relax PMLR, 2023, pp. 14\,023--14\,038.

\bibitem{RPCANet}
F.~Wu, T.~Zhang, L.~Li, Y.~Huang, and Z.~Peng, ``Rpcanet: Deep unfolding rpca
  based infrared small target detection,'' in \emph{Proceedings of the IEEE/CVF
  Winter Conference on Applications of Computer Vision (WACV)}, January 2024,
  pp. 4809--4818.

\bibitem{ding2022scaling}
X.~Ding, X.~Zhang, J.~Han, and G.~Ding, ``Scaling up your kernels to 31x31:
  Revisiting large kernel design in cnns,'' in \emph{Proceedings of the
  IEEE/CVF conference on computer vision and pattern recognition}, 2022, pp.
  11\,963--11\,975.

\bibitem{ding2024unireplknet}
X.~Ding, Y.~Zhang, Y.~Ge, S.~Zhao, L.~Song, X.~Yue, and Y.~Shan, ``Unireplknet:
  A universal perception large-kernel convnet for audio video point cloud
  time-series and image recognition,'' in \emph{Proceedings of the IEEE/CVF
  Conference on Computer Vision and Pattern Recognition}, 2024, pp. 5513--5524.

\bibitem{hou2024conv2former}
Q.~Hou, C.-Z. Lu, M.-M. Cheng, and J.~Feng, ``Conv2former: A simple
  transformer-style convnet for visual recognition,'' \emph{IEEE Transactions
  on Pattern Analysis and Machine Intelligence}, 2024.

\bibitem{li2023moganet}
S.~Li, Z.~Wang, Z.~Liu, C.~Tan, H.~Lin, D.~Wu, Z.~Chen, J.~Zheng, and S.~Z. Li,
  ``Moganet: Multi-order gated aggregation network,'' in \emph{The Twelfth
  International Conference on Learning Representations}, 2023.

\bibitem{finder2025wavelet}
S.~E. Finder, R.~Amoyal, E.~Treister, and O.~Freifeld, ``Wavelet convolutions
  for large receptive fields,'' in \emph{European Conference on Computer
  Vision}.\hskip 1em plus 0.5em minus 0.4em\relax Springer, 2025, pp. 363--380.

\bibitem{lau2024large}
K.~W. Lau, L.-M. Po, and Y.~A.~U. Rehman, ``Large separable kernel attention:
  Rethinking the large kernel attention design in cnn,'' \emph{Expert Systems
  with Applications}, vol. 236, p. 121352, 2024.

\bibitem{chen2024pelk}
H.~Chen, X.~Chu, Y.~Ren, X.~Zhao, and K.~Huang, ``Pelk: Parameter-efficient
  large kernel convnets with peripheral convolution,'' in \emph{Proceedings of
  the IEEE/CVF Conference on Computer Vision and Pattern Recognition}, 2024,
  pp. 5557--5567.

\bibitem{guo2022segnext}
M.-H. Guo, C.-Z. Lu, Q.~Hou, Z.~Liu, M.-M. Cheng, and S.-M. Hu, ``Segnext:
  Rethinking convolutional attention design for semantic segmentation,''
  \emph{Advances in Neural Information Processing Systems}, vol.~35, pp.
  1140--1156, 2022.

\bibitem{li2024lsknet}
Y.~Li, X.~Li, Y.~Dai, Q.~Hou, L.~Liu, Y.~Liu, M.-M. Cheng, and J.~Yang,
  ``Lsknet: A foundation lightweight backbone for remote sensing,''
  \emph{International Journal of Computer Vision}, pp. 1--22, 2024.

\bibitem{li2023large}
Y.~Li, Q.~Hou, Z.~Zheng, M.-M. Cheng, J.~Yang, and X.~Li, ``Large selective
  kernel network for remote sensing object detection,'' in \emph{Proceedings of
  the IEEE/CVF International Conference on Computer Vision}, 2023, pp.
  16\,794--16\,805.

\bibitem{azad2024beyond}
R.~Azad, L.~Niggemeier, M.~H{\"u}ttemann, A.~Kazerouni, E.~K. Aghdam,
  Y.~Velichko, U.~Bagci, and D.~Merhof, ``Beyond self-attention: Deformable
  large kernel attention for medical image segmentation,'' in \emph{Proceedings
  of the IEEE/CVF Winter Conference on Applications of Computer Vision}, 2024,
  pp. 1287--1297.

\bibitem{rao2022hornet}
Y.~Rao, W.~Zhao, Y.~Tang, J.~Zhou, S.~N. Lim, and J.~Lu, ``Hornet: Efficient
  high-order spatial interactions with recursive gated convolutions,''
  \emph{Advances in Neural Information Processing Systems}, vol.~35, pp.
  10\,353--10\,366, 2022.

\bibitem{ho2020denoising}
J.~Ho, A.~Jain, and P.~Abbeel, ``Denoising diffusion probabilistic models,''
  \emph{Advances in neural information processing systems}, vol.~33, pp.
  6840--6851, 2020.

\bibitem{nichol2021improved}
A.~Q. Nichol and P.~Dhariwal, ``Improved denoising diffusion probabilistic
  models,'' in \emph{International conference on machine learning}.\hskip 1em
  plus 0.5em minus 0.4em\relax PMLR, 2021, pp. 8162--8171.

\bibitem{lugmayr2022repaint}
A.~Lugmayr, M.~Danelljan, A.~Romero, F.~Yu, R.~Timofte, and L.~Van~Gool,
  ``Repaint: Inpainting using denoising diffusion probabilistic models,'' in
  \emph{Proceedings of the IEEE/CVF conference on computer vision and pattern
  recognition}, 2022, pp. 11\,461--11\,471.

\bibitem{yu2022metaformer}
W.~Yu, M.~Luo, P.~Zhou, C.~Si, Y.~Zhou, X.~Wang, J.~Feng, and S.~Yan,
  ``Metaformer is actually what you need for vision,'' in \emph{Proceedings of
  the IEEE/CVF conference on computer vision and pattern recognition}, 2022,
  pp. 10\,819--10\,829.

\bibitem{yu2023metaformer}
W.~Yu, C.~Si, P.~Zhou, M.~Luo, Y.~Zhou, J.~Feng, S.~Yan, and X.~Wang,
  ``Metaformer baselines for vision,'' \emph{IEEE Transactions on Pattern
  Analysis and Machine Intelligence}, 2023.

\bibitem{shi2024transnext}
D.~Shi, ``Transnext: Robust foveal visual perception for vision transformers,''
  in \emph{Proceedings of the IEEE/CVF Conference on Computer Vision and
  Pattern Recognition}, 2024, pp. 17\,773--17\,783.

\bibitem{10595430}
H.~Wu, C.~Wang, L.~Tu, C.~Patsch, and Z.~Jin, ``Cspn: A category-specific
  processing network for low-light image enhancement,'' \emph{IEEE Transactions
  on Circuits and Systems for Video Technology}, vol.~34, no.~11, pp.
  11\,929--11\,941, 2024.

\bibitem{huang2023revisiting}
S.~Huang, Z.~Lu, K.~Deb, and V.~N. Boddeti, ``Revisiting residual networks for
  adversarial robustness,'' in \emph{Proceedings of the IEEE/CVF Conference on
  Computer Vision and Pattern Recognition}, 2023, pp. 8202--8211.

\bibitem{he2019mgnet}
J.~He and J.~Xu, ``Mgnet: A unified framework of multigrid and convolutional
  neural network,'' \emph{Science china mathematics}, vol.~62, pp. 1331--1354,
  2019.

\bibitem{wu2018group}
Y.~Wu and K.~He, ``Group normalization,'' in \emph{Proceedings of the European
  conference on computer vision (ECCV)}, 2018, pp. 3--19.

\bibitem{touvron2021going}
H.~Touvron, M.~Cord, A.~Sablayrolles, G.~Synnaeve, and H.~J{\'e}gou, ``Going
  deeper with image transformers,'' in \emph{Proceedings of the IEEE/CVF
  international conference on computer vision}, 2021, pp. 32--42.

\bibitem{xie2021segformer}
E.~Xie, W.~Wang, Z.~Yu, A.~Anandkumar, J.~M. Alvarez, and P.~Luo, ``Segformer:
  Simple and efficient design for semantic segmentation with transformers,''
  \emph{Advances in neural information processing systems}, vol.~34, pp.
  12\,077--12\,090, 2021.

\bibitem{wang2021pyramid}
W.~Wang, E.~Xie, X.~Li, D.-P. Fan, K.~Song, D.~Liang, T.~Lu, P.~Luo, and
  L.~Shao, ``Pyramid vision transformer: A versatile backbone for dense
  prediction without convolutions,'' in \emph{Proceedings of the IEEE/CVF
  international conference on computer vision}, 2021, pp. 568--578.

\bibitem{he2016deep}
K.~He, X.~Zhang, S.~Ren, and J.~Sun, ``Deep residual learning for image
  recognition,'' in \emph{Proceedings of the IEEE conference on computer vision
  and pattern recognition}, 2016, pp. 770--778.

\bibitem{zhang2022resnest}
H.~Zhang, C.~Wu, Z.~Zhang, Y.~Zhu, H.~Lin, Z.~Zhang, Y.~Sun, T.~He, J.~Mueller,
  R.~Manmatha \emph{et~al.}, ``Resnest: Split-attention networks,'' in
  \emph{Proceedings of the IEEE/CVF conference on computer vision and pattern
  recognition}, 2022, pp. 2736--2746.

\bibitem{hendrycks2016gaussian}
D.~Hendrycks and K.~Gimpel, ``Gaussian error linear units (gelus),''
  \emph{arXiv preprint arXiv:1606.08415}, 2016.

\bibitem{chen2023run}
J.~Chen, S.-h. Kao, H.~He, W.~Zhuo, S.~Wen, C.-H. Lee, and S.-H.~G. Chan,
  ``Run, don't walk: chasing higher flops for faster neural networks,'' in
  \emph{Proceedings of the IEEE/CVF conference on computer vision and pattern
  recognition}, 2023, pp. 12\,021--12\,031.

\bibitem{li2019selective}
X.~Li, W.~Wang, X.~Hu, and J.~Yang, ``Selective kernel networks,'' in
  \emph{Proceedings of the IEEE/CVF conference on computer vision and pattern
  recognition}, 2019, pp. 510--519.

\bibitem{hinton2015distilling}
G.~Hinton, ``Distilling the knowledge in a neural network,'' \emph{arXiv
  preprint arXiv:1503.02531}, 2015.

\bibitem{xu2024eviprompt}
Y.~Xu, J.~Tang, A.~Men, and Q.~Chen, ``Eviprompt: A training-free evidential
  prompt generation method for adapting segment anything model in medical
  images,'' \emph{IEEE Transactions on Image Processing}, 2024.

\bibitem{ma2024rewrite}
X.~Ma, X.~Dai, Y.~Bai, Y.~Wang, and Y.~Fu, ``Rewrite the stars,'' in
  \emph{Proceedings of the IEEE/CVF Conference on Computer Vision and Pattern
  Recognition}, 2024, pp. 5694--5703.

\bibitem{wang2024multi}
Y.~Wang, Y.~Li, G.~Wang, and X.~Liu, ``Multi-scale attention network for single
  image super-resolution,'' in \emph{Proceedings of the IEEE/CVF Conference on
  Computer Vision and Pattern Recognition}, 2024, pp. 5950--5960.

\bibitem{MRF3Net}
X.~Zhang, X.~Zhang, S.-Y. Cao, B.~Yu, C.~Zhang, and H.-L. Shen, ``Mrf3net: An
  infrared small target detection network using multireceptive field perception
  and effective feature fusion,'' \emph{IEEE Transactions on Geoscience and
  Remote Sensing}, vol.~62, pp. 1--14, 2024.

\bibitem{xie2023revealing}
Z.~Xie, Z.~Geng, J.~Hu, Z.~Zhang, H.~Hu, and Y.~Cao, ``Revealing the dark
  secrets of masked image modeling,'' in \emph{Proceedings of the IEEE/CVF
  conference on computer vision and pattern recognition}, 2023, pp.
  14\,475--14\,485.

\bibitem{SoftIoU}
Y.~Huang, Z.~Tang, D.~Chen, K.~Su, and C.~Chen, ``Batching soft iou for
  training semantic segmentation networks,'' \emph{IEEE Signal Processing
  Letters}, vol.~27, pp. 66--70, 2020.

\bibitem{Cheng_2021_CVPR}
B.~Cheng, R.~Girshick, P.~Dollar, A.~C. Berg, and A.~Kirillov, ``Boundary iou:
  Improving object-centric image segmentation evaluation,'' in
  \emph{Proceedings of the IEEE/CVF Conference on Computer Vision and Pattern
  Recognition (CVPR)}, June 2021, pp. 15\,334--15\,342.

\bibitem{ADAN}
X.~Xie, P.~Zhou, H.~Li, Z.~Lin, and S.~Yan, ``Adan: Adaptive nesterov momentum
  algorithm for faster optimizing deep models,'' \emph{IEEE Transactions on
  Pattern Analysis and Machine Intelligence}, vol.~46, no.~12, pp. 9508--9520,
  2024.

\bibitem{wang2019miss}
H.~Wang, L.~Zhou, and L.~Wang, ``Miss detection vs. false alarm: Adversarial
  learning for small object segmentation in infrared images,'' in
  \emph{Proceedings of the IEEE/CVF International Conference on Computer
  Vision}, 2019, pp. 8509--8518.

\bibitem{FC3-Net}
M.~Zhang, K.~Yue, J.~Zhang, Y.~Li, and X.~Gao, ``Exploring feature compensation
  and cross-level correlation for infrared small target detection,'' in
  \emph{Proceedings of the 30th ACM International Conference on Multimedia},
  ser. MM '22.\hskip 1em plus 0.5em minus 0.4em\relax New York, NY, USA:
  Association for Computing Machinery, 2022, p. 1857–1865.

\bibitem{liu2022convnet}
Z.~Liu, H.~Mao, C.-Y. Wu, C.~Feichtenhofer, T.~Darrell, and S.~Xie, ``A convnet
  for the 2020s,'' in \emph{Proceedings of the IEEE/CVF conference on computer
  vision and pattern recognition}, 2022, pp. 11\,976--11\,986.

\bibitem{shleifer2021normformer}
S.~Shleifer, J.~Weston, and M.~Ott, ``Normformer: Improved transformer
  pretraining with extra normalization,'' \emph{arXiv preprint
  arXiv:2110.09456}, 2021.

\bibitem{ma2018shufflenet}
N.~Ma, X.~Zhang, H.-T. Zheng, and J.~Sun, ``Shufflenet v2: Practical guidelines
  for efficient cnn architecture design,'' in \emph{Proceedings of the European
  conference on computer vision (ECCV)}, 2018, pp. 116--131.

\end{thebibliography}

\newpage

\vfill

\end{document}